\documentclass[10pt,journal,compsoc]{IEEEtran}

\usepackage{times}
\usepackage{epsfig}
\usepackage{graphicx}
\usepackage{amsmath}
\usepackage{amssymb}
\usepackage{multirow}
\usepackage{textcomp}
\usepackage[utf8]{inputenc}
\usepackage{enumitem}
\usepackage{textcomp}
\usepackage{xcolor}

\begin{document}

\title{Analysis of the hands in egocentric vision:\\ A survey}

\author{Andrea~Bandini,~\IEEEmembership{Member,~IEEE,}
        and~José~Zariffa,~\IEEEmembership{Senior~Member,~IEEE}
\IEEEcompsocitemizethanks{\IEEEcompsocthanksitem A. Bandini and J. Zariffa are with KITE – Toronto Rehab – University
Health Network, Toronto, ON, CA.\protect\\
\IEEEcompsocthanksitem J. Zariffa is also with the Institute of Biomaterials and
Biomedical Engineering (IBBME), University of Toronto, Toronto, ON, CA, the Edward S. Rogers Sr. Department of Electrical and
Computer Engineering, University of Toronto, Toronto, ON, CA, and the Rehabilitation Sciences Institute, University of Toronto, Toronto, ON, CA.\protect\\
E-mail: andrea.bandini@uhn.ca}}
\IEEEtitleabstractindextext{%
\begin{abstract}
Egocentric vision (a.k.a. first-person vision -- FPV) applications have thrived over the past few years, thanks to the availability of affordable wearable cameras and large annotated datasets. The position of the wearable camera (usually mounted on the head) allows recording exactly what the camera wearers have in front of them, in particular hands and manipulated objects. This intrinsic advantage enables the study of the hands from multiple perspectives: localizing hands and their parts within the images; understanding what actions and activities the hands are involved in; and developing human-computer interfaces that rely on hand gestures. In this survey, we review the literature that focuses on the hands using egocentric vision, categorizing the existing approaches into: localization (where are the hands or parts of them?); interpretation (what are the hands doing?); and application (e.g., systems that used egocentric hand cues for solving a specific problem). Moreover, a list of the most prominent datasets with hand-based annotations is provided.
\end{abstract}

}

\maketitle
\IEEEdisplaynontitleabstractindextext

%
\IEEEpeerreviewmaketitle


\IEEEraisesectionheading{\section{Introduction}\label{sec:introduction}}
\IEEEPARstart{T}{he} hands are of primary importance for human beings, as they allow us to interact with objects and environments, communicate with other people, and perform activities of daily living (ADLs) such as eating, bathing, and dressing. It is not a surprise that in individuals with impaired or reduced hand functionality (e.g., after a stroke or cervical spinal cord injury -- cSCI) the top recovery priority is to regain the function of the hands \cite{snoek2004survey}. Given their importance, computer vision researchers have tried to analyze the hands from multiple perspectives: localizing them in the images \cite{li2013pixel}, inferring the types of actions they are involved in \cite{fathi2011understanding}, as well as enabling interactions with computers and robots \cite{rautaray2015vision,zabulis2009vision}. Wearable cameras (e.g., cameras mounted on the head or chest) have allowed studying the hands from a point of view (POV) that provides a first-person perspective of the world. This field of research in computer vision is known as egocentric or first-person vision (FPV). Although some studies were published as early as the 1990s \cite{mann1998wearcam}, FPV gained more importance after 2012 with the emergence of smart glasses and action cameras (i.e., Google Glass and GoPro cameras). For an overview of the evolution of FPV methods, the reader is referred to the survey published by Betancourt et al. \cite{betancourt2015evolution}.

Egocentric vision presents many advantages when compared with third person vision, where the camera position is usually stable and disjointed from the user. For example: the device is recording exactly what the users have in front of them; camera movement is driven by the camera-wearer’s activity and attention; hands and manipulated objects tend to appear at the center of the image and hand occlusions are minimized \cite{nguyen2016recognition}. These advantages made the development of novel approaches for studying the hands very appealing. However, when working in FPV, researchers must also face an important issue: the camera is not stable, but is moving with the human body. This causes fast movements and sudden illumination changes that can significantly reduce the quality of the video recordings and make it more difficult to separate the hand and objects of interest from the background.

Betancourt et al. \cite{betancourt2015towards} clearly summarized the typical processing steps of hand-based methods in FPV. The authors proposed a unified and hierarchical framework where the lowest levels of the hierarchy concern the detection and segmentation of the hands, whereas the highest levels are related to interaction and activity recognition. Each level is devoted to a specific task and provides the results to higher levels (e.g., hand identification builds upon hand segmentation and hand detection, activity recognition builds upon the identification of interactions, etc.). Although clear and concise, this framework could not cover some of the recent developments in this field, made possible thanks to the availability of large amounts of annotated data and to the advent of deep learning \cite{zhang2018egogesture,wang2019recurrent,garcia2018first}. Other good surveys closely related to the topics discussed in our paper were published in the past few years \cite{rautaray2015vision,nguyen2016recognition,cheng2015survey,li2019survey,del2016summarization,bolanos2016toward}. The reader should refer to the work of Del Molino et al. \cite{del2016summarization} for an introduction into video summarization in FPV, to the survey of Nguyen et al. \cite{nguyen2016recognition} for the recognition of ADLs from egocentric vision, and to the work of Bola\~{n}os et al. \cite{bolanos2016toward} for a review on visual lifelogging. Hand pose estimation and hand gesture recognition methods are analyzed in \cite{li2019survey} and \cite{cheng2015survey}, respectively.

In this survey we define a comprehensive taxonomy of hand-based methods in FPV expanding the categorization proposed in \cite{betancourt2015towards} and classifying the existing literature into three macro-areas: localization, interpretation, and application. For each macro-area we identify the main sub-areas of research, presenting the most prominent approaches published in the past 10 years and discussing advantages and disadvantages of each method. Moreover, we summarize the available datasets published in this field. Our focus in defining a comprehensive taxonomy and comparing different approaches is to propose an updated and general framework of hand-based methods in FPV, highlighting the current trends and summarizing the main findings, in order to provide guidelines to researchers who want to improve and expand this field of research.
The remainder of the paper is organized as follows: Section 2 presents a taxonomy of hand-based methods in FPV following a novel categorization that divides these approaches into three macro-areas: localization, interpretation, and application; Section 3 describes the approaches developed for solving the localization problem; Section 4 summarizes the work focused on interpretation; Section 5 summarizes the most important applications of hand-based methods in FPV; Section 6 reviews the available datasets published so far; and, finally, Section 7 concludes with a discussion of the current trends in this field.


\section{Hand-based methods in FPV -- An updated framework}\label{sec:framework}
Starting from the raw frames, the first processing step is dedicated to the localization of the hands or parts of them within the observed scene. This allows restricting the processing to small regions of interest (ROIs), excluding unnecessary information from the background, or reducing the dimensionality of the problem, by extracting the articulated hand pose. Once the positions of the hands and/or their joints have been determined, higher-level information can be inferred to understand what the hands are doing (e.g. gesture and posture recognition, action and activity recognition). This information can be used for building applications such as human-computer interaction (HCI) and human-robot interaction (HRI) \cite{rautaray2015vision,zabulis2009vision}. Therefore, we categorize the existing studies that made use of hand-based methods in FPV into three macro-areas:
\begin{itemize}[noitemsep, wide=0pt, leftmargin=\dimexpr\labelwidth + 2\labelsep\relax]
    \item \textbf{Localization} -- approaches that answer the question: \textbf{where} are the hands (or parts of them)?
    \item \textbf{Interpretation} -- approaches that answer the question: \textbf{what} are the hands doing?
    \item \textbf{Application} -- approaches that use methods from the above areas to build real-world applications.
\end{itemize}
For each area we define sub-areas according to the aims and nature of the proposed methods.

\subsection{Localization -- Where are the hands (or parts of them)?}
\begin{figure}
\begin{center}
\includegraphics[width=0.95\linewidth]{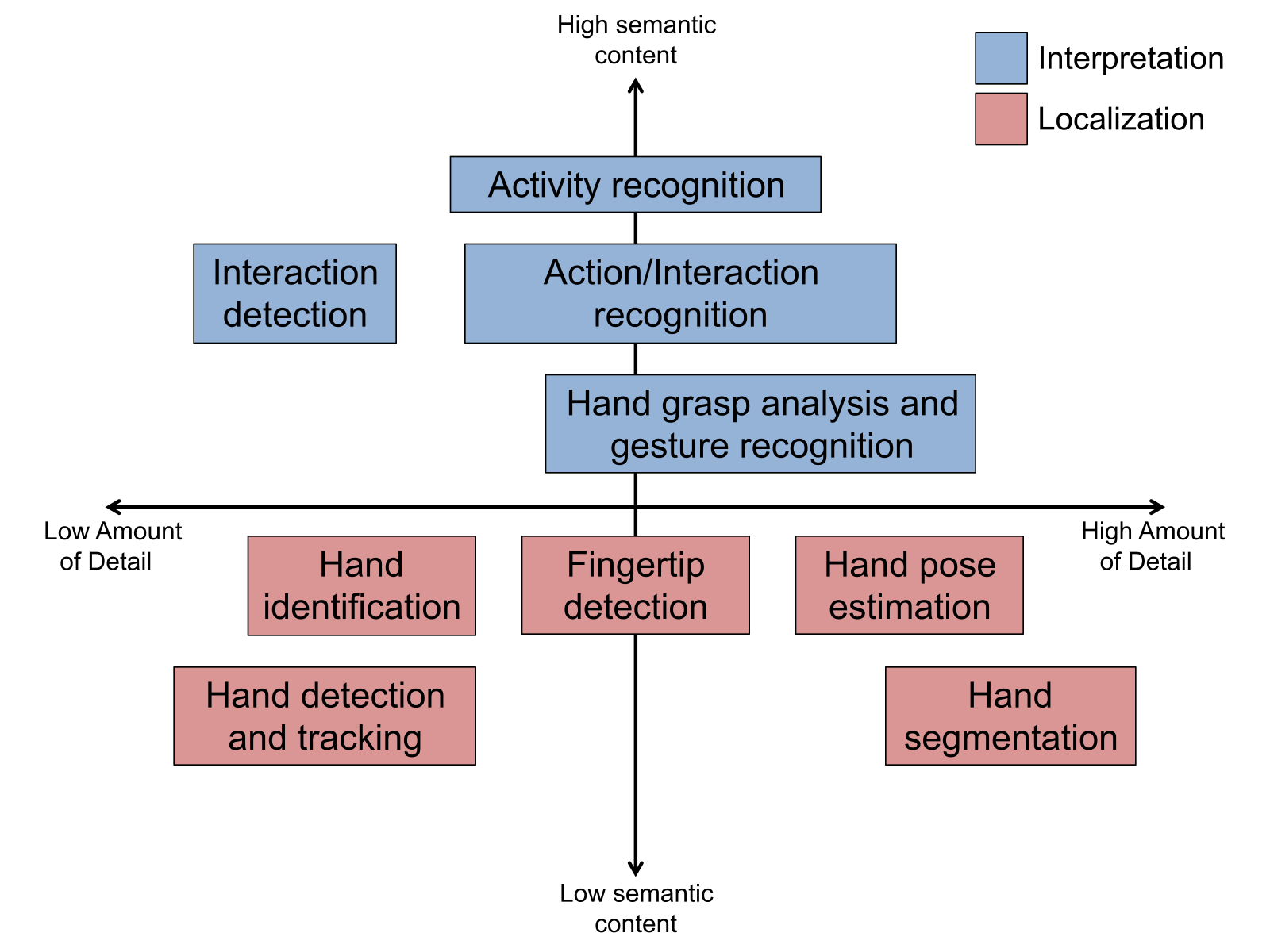}
\end{center}
\caption{Hand-based approaches in FPV categorized by amount of detail and semantic content.}
\label{fig_framework}
\end{figure}

The localization area encloses all the approaches that aim at localizing hands (or parts of them) within the images. The sub-areas are:
\begin{itemize}[noitemsep, wide=0pt, leftmargin=\dimexpr\labelwidth + 2\labelsep\relax]
    \item \textbf{Hand segmentation} -- detecting the hand regions with pixel-level detail.
    \item \textbf{Hand detection} -- defined both as binary classification problem (does the image contain a hand?) and object localization problem (is there a hand? Where is it located?). The generalization of hand detection over time is \textbf{hand tracking}.
    \item \textbf{Hand identification} -- classification between left and right hand, as well as other hands present in the scene. 
    \item \textbf{Hand pose estimation} -- estimation of hand joint positions. A simplified version of the hand pose estimation problem is \textbf{fingertip detection}, where only the fingertips of one or more fingers are identified.
\end{itemize}
From the above sub-areas it is possible to highlight two dimensions in the localization problem. The first one is the amount of detail of the information extracted with a method. For example, hand detection results in low-detail information (i.e., binary label or coordinates of a bounding box), whereas hand segmentation produces high-detail information (i.e., pixel-level silhouette). The second dimension is the meaning of the obtained information \cite{nguyen2016recognition,chaaraoui2012review}, hereafter called semantic content. Hand detection and segmentation, although producing different amounts of detail, have the same semantic content, namely the global position of the hand. By contrast, hand pose estimation has higher semantic content than hand detection, as the position of the fingers and hand joints add more information to the global hand location. This categorization is shown in figure \ref{fig_framework}.

\subsection{Interpretation -- What are the hands doing?}
The interpretation area includes those approaches that, starting from lower level information (i.e., detection, segmentation, pose estimation, etc.), try to infer information with higher semantic content. The main sub-areas are:
\begin{itemize}[noitemsep, wide=0pt, leftmargin=\dimexpr\labelwidth + 2\labelsep\relax]
    \item \textbf{Hand grasp analysis} -- Detection of the dominant hand postures during hand-object interactions.
    \item \textbf{Hand gesture recognition} -- Classification of hand gestures, usually as input for virtual reality (VR) and augmented reality (AR) systems, as will be discussed in Section \ref{sec:application}.
    \item \textbf{Action/Interaction recognition} -- Predicting what type of action or interaction the hands are involved in. Following the taxonomy of Tekin et al. \cite{tekin2019h+}, an action is defined as a verb (e.g. “pour”), whereas an interaction as a verb-noun pair (e.g. “pour water”). This task is called \textit{interaction detection} if the problem is reduced to a binary classification task (i.e., predicting whether or not the hands are interacting).
    \item \textbf{Activity recognition} -- Identification of the activities, defined as set of temporally-consistent actions \cite{fathi2011understanding}. For example, preparing a meal is an activity composed of several actions and interactions, such as cutting vegetables, pouring water, opening jars, etc.
\end{itemize}
We can qualitatively compare these sub-areas according to the two dimensions described above (i.e., amount of detail and semantic content). Hand grasp analysis and gesture recognition have lower semantic content than action/interaction recognition that, in turn has lower semantic content than activity recognition. Activity recognition, although with higher semantic content than action recognition, produces results with lower detail. This is because the information is summarized towards the upper end of the semantic content dimension.
Following these considerations, we represent the localization and interpretation areas of this framework on a two-dimensional plot whose axes are the amount of detail and the semantic content (see Figure \ref{fig_framework}).

\subsection{Application}
The application area includes all the FPV approaches and systems that make use of hand-based methods for achieving certain objectives. The main applications are:
\begin{itemize}[noitemsep, wide=0pt, leftmargin=\dimexpr\labelwidth + 2\labelsep\relax]
    \item Healthcare application, for example the remote assessment of hand function and the development of ambient assisted living (AAL) systems.
    \item HCI and HRI, for example VR and AR applications, or HRI systems that rely on the recognition of hand gestures.
\end{itemize}
Some egocentric vision applications were already covered by other surveys \cite{nguyen2016recognition,del2016summarization, bolanos2016toward, rautaray2015vision}. Thus, we will summarizing novel aspects related to hand-based methods in FPV not covered in the previous articles.


\section{Localization}\label{sec:localization}
The localization of hands (or parts of them) is the first and most important processing step of many hand-based methods in FPV. A good hand localization algorithm allows estimating the accurate position of the hands within the image, boosting the performance of higher-level inference \cite{bambach2015lending}. For this reason, hand localization has been the main focus of researchers in egocentric vision. Although many hand detection, pose-estimation, and segmentation algorithms were developed in third person vision \cite{li2019survey}, the egocentric POV presents notable challenges that do not allow a direct translation of these methods. Rogez et al. \cite{rogez20143d} demonstrated that egocentric hand detection is considerably harder in FPV, and methods developed specifically for third person POV may fail when applied to egocentric videos.

Hand segmentation and detection are certainly the two most extensively studied sub-areas. They are often used in combination, for example to classify as “hand” or “not hand” previously segmented regions \cite{zhu2016two,cartas2017detecting}, or to segment ROIs previously obtained with a hand detector \cite{bambach2015lending}. However, considering the extensive research behind these two sub-areas, we summarize them separately.

\begin{figure*}[t]
\begin{center}
\includegraphics[width=0.9\linewidth]{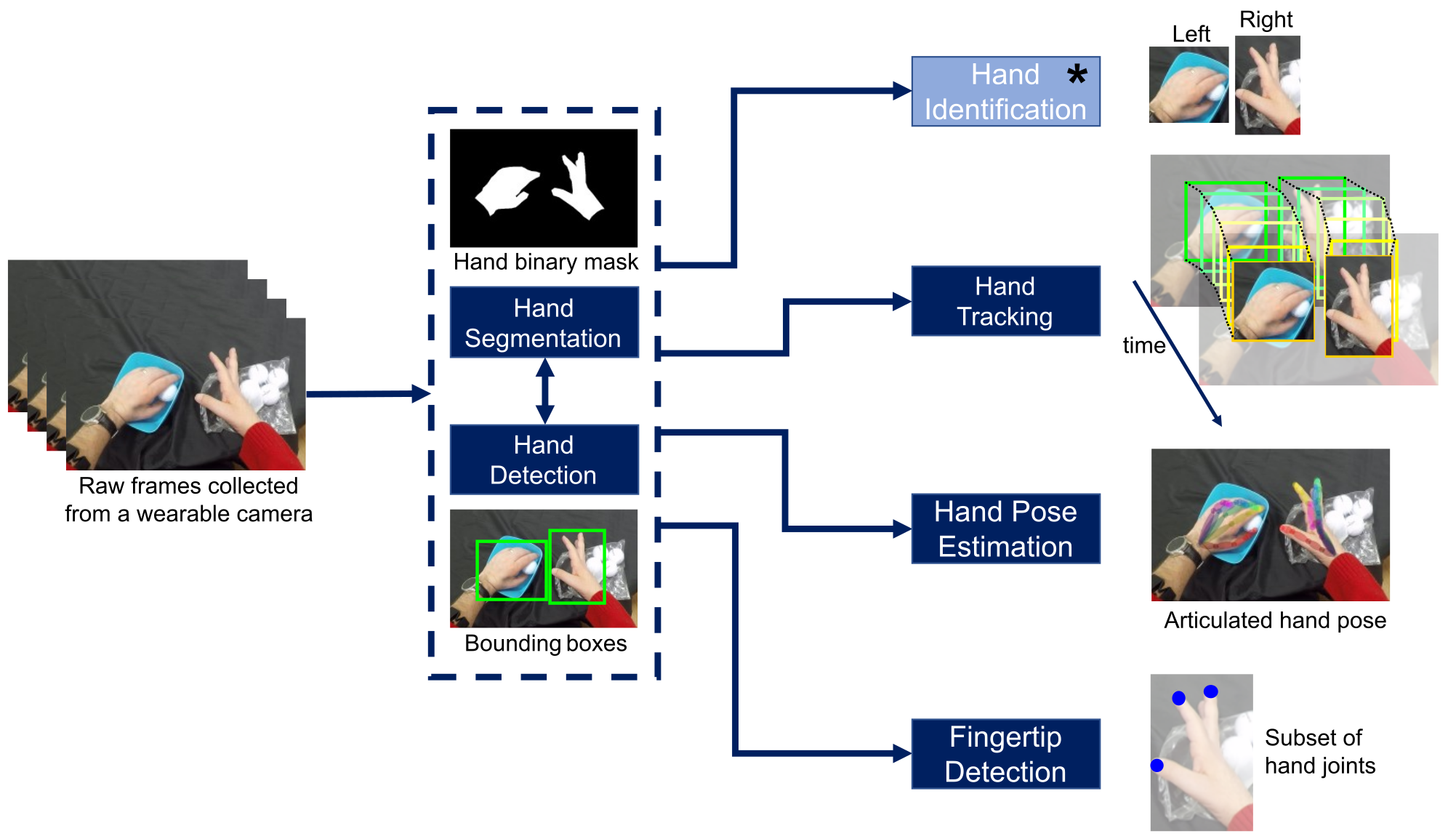}
\end{center}
\caption{Diagram of hand localization tasks in egocentric vision. Hand detection and segmentation have often been used in combination, for example to segment ROIs previously obtained with a hand detector, or to classify as “hand” or “not hand” previously segmented regions. Since they provide the global position of the hands within the frame, they are chosen as the basis for other localization approaches, such as hand identification, hand tracking, hand pose estimation, fingertip detection (\textbf{*}: Hand identification is now typically incorporated within the hand detection step).}
\label{fig_localization}
\end{figure*}

\subsection{Hand segmentation}\label{subsec:handSeg}
Hand segmentation is the process of identifying the hand regions at pixel-level (see Figure \ref{fig_localization}). This step allows extracting the silhouette of the hands and has been extensively used as a pre-processing step for hand pose estimation, hand-gesture recognition, action/interaction recognition, and activity recognition. One of the most straightforward approaches is to use the color as discriminative feature to identify skin-like pixels \cite{jones2002statistical}. Although very simple and fast, color-based segmentation fails whenever background objects have similar skin color (e.g., wooden objects) and it is robust only if the user wears colored gloves or patches to simplify the processing \cite{liang2015egocentric,likitlersuang2016interaction}. However, this might not be feasible in real-world applications, where the hand segmentation algorithm is supposed to work without external cues, thus mimicking human vision. Illumination changes due to different environments also negatively affect the segmentation performance. Moreover, the availability of large datasets with pixel-level ground truth annotations is another issue when working with hand segmentation. This type of annotation requires a lot of manual work and the size of these datasets is much smaller than those with less detailed annotations (e.g., bounding boxes). Thus, several approaches were proposed to face the above issues.

\subsubsection{Discriminating hands from objects and background}\label{subsubsec:segHandsObjBack}
Traditional hand segmentation approaches (i.e., not based on deep learning) rely on the extraction of features from an image patch, classifying the central pixel or the entire patch as skin or no-skin using a binary classifier or regression model. The vast majority of approaches combined color with gradient and/or texture features, whereas random forest has been the most popular classification algorithm \cite{breiman2001random}. The use of texture and gradient features allows capturing salient patterns and contours of the hands that, combined with the color features, help discriminate them from background and objects with similar color.

\textbf{Pixel-based classification}. Li and Kitani \cite{li2013pixel} tested different combinations of color (HSV, RGB, and LAB color spaces) and local appearance features (Gabor filters \cite{weldon1996efficient}, HOG \cite{dalal2005histograms}, SIFT \cite{lowe1999object}, BRIEF \cite{calonder2010brief}, and ORB \cite{rublee2011orb} descriptors) to capture local contours and gradients of the hand regions. Each pixel was classified as skin or no-skin using a random forest regression. When using color features alone, the LAB color space provided the best performance, whereas gradient and texture features, such as HOG and BRIEF, improved the segmentation performance when combined with the color information \cite{li2013pixel}. Zariffa and Popovic \cite{zariffa2013hand} used a mixture of Gaussian skin model with dilation and erosion morphological operators to detect a coarse estimate of the hand regions. The initial region was refined by removing small isolated blobs with texture different from the skin, by computing the Laplacian of the image within each blob. Lastly, pixel-level segmentation was achieved by backprojecting using an adaptively selected region in the colour space. In \cite{likitlersuang2019egocentric}, the coarse segmentation obtained with a mixture of Gaussian skin model \cite{jones2002statistical,zariffa2013hand} was refined by using a structured forest edge detection \cite{dollar2013structured}, specifically trained on available datasets \cite{cartas2017detecting,betancourt2014sequential}. 

\textbf{Patch-based classification}. Other authors classified image patches instead of single pixels, in order to produce segmentation masks more robust to pixel-level noise \cite{serra2013hand,singh2016first,urabe2018cooking,zhu2014pixel,zhu2015structured}. Serra et al. \cite{serra2013hand} classified clusters of pixels (i.e., super-pixels) obtained with the simple linear iterative clustering (SLIC) algorithm \cite{achanta2012slic}. For each super-pixel, they used a combination of color (HSV and LAB color spaces) and gradient features (Gabor filters and histogram of gradients) to train a random forest classifier. The segmentation was refined by assuming temporal coherence between consecutive frames and spatial consistency among groups of super-pixels. Similarly, Singh et al. \cite{singh2016first} computed the hand binary mask by extracting texture and color features (Gabor filters with RGB, HSV, and LAB color features) from the super-pixels, whereas Urabe et al. \cite{urabe2018cooking} used the same features in conjunction with the centroid location of each super-pixel to train a support vector machine (SVM) for segmenting the skin regions. Instead of classifying the whole patch from which color, gradient and texture features are extracted, Zhu et al. \cite{zhu2014pixel,zhu2015structured} learned the segmentation mask within the image patch, by using a random forest framework (i.e., shape-aware structured forest).

\textbf{Deep learning} may help solve hand segmentation problems in FPV. However, its use is still hampered by the lack of large annotated datasets with pixel-level annotations. Some deep learning approaches for hand segmentation \cite{zhou2016cascaded,li2019supervised} tackled this issue by using the available annotations in combination with other image segmentation techniques (e.g., super-pixels or GrabCut \cite{achanta2012slic,li2015superpixel,felzenszwalb2004efficient,rother2004grabcut}) to generate new hand segmentation masks for expanding the dataset and fine-tuning pre-trained networks (see Section \ref{subsubsec:segLackAnnot} for more details). The availability of pre-trained convolutional neural networks (CNNs) for semantic object segmentation \cite{long2015fully,lin2017refinenet} was exploited in \cite{urooj2018analysis,tang2018multi}. Wang et al. \cite{wang2018beyond,wang2019recurrent} tackled the hand segmentation problem in a recurrent manner by using a recurrent U-NET architecture \cite{ronneberger2015u}. The rationale behind this strategy is to imitate the saccadic movements of the eyes that allow refining the perception of a scene. The computational cost can be another issue in CNN-based hand segmentation. To reduce this cost, while achieving good segmentation accuracy, Li et al. \cite{li2019flow} implemented the deep feature flow (DFF) \cite{zhu2017deep} with an extra branch to make the approach more robust against occlusions and distortion caused by DFF.

\subsubsection{Robustness to illumination changes}\label{subsubsec:segIllChange}
The problem of variable illumination can be partially alleviated by choosing the right color-space for feature extraction (e.g., LAB \cite{li2013pixel}) and increasing the size of the training set. However, the latter strategy may reduce the separability of the color space and increase the number misclassified examples \cite{betancourt2016gpu}. Thus, a popular solution has been to use a collection of segmentation models, adaptively selecting the most appropriate one for the current test conditions \cite{li2013pixel,serra2013hand,singh2016first,betancourt2016gpu,li2013model}. Li and Kitani \cite{li2013pixel} proposed an adaptive approach that selects the nearest segmentation model, namely the one trained in a similar environment. To learn different global appearance models, they clustered the HSV histogram of the training images using k-means and learned a separate random tree regressor for each cluster. They further extended this concept in \cite{li2013model} where they formulated hand segmentation as a model recommendation task. For each test image, the system was able to propose the best hand segmentation model given the color and structure (HSV histogram and HOG features) of the observed scene and the relative performance between two segmentation models. Similarly, Betancourt et al. \cite{betancourt2016gpu} trained binary random forests to classify each pixel as skin or not skin using the LAB values. For each frame they trained a separate segmentation model storing it along with the HSV histogram, as a proxy to summarize the illumination condition of that frame. K-nearest neighbors (k-NN) classification was performed on the global features to select the \textit{k} most suitable segmentation models. These models were applied to the test frame and their segmentation results combined together to obtain the final hand mask. 

\subsubsection{Lack of pixel-level annotations}\label{subsubsec:segLackAnnot}
Annotating images at pixel-level is a very laborious and costly work that refrains many authors from publishing large annotated datasets. Thus, the ideal solution to the hand segmentation problem would be a self-supervised approach able to learn the appearance of the hands on-the-fly, or a weakly supervised method that relies on the available training data to produce new hand masks.

Usually, methods for online hand segmentation made assumptions on the hand motion \cite{huang2016egocentric,huang2018egocentric,zhao2017unsupervised,zhao2018coarse} and/or required the user to perform a calibration with pre-defined hand movements \cite{kumar2015fly}. In this way, the combination of color and motion features facilitates the detection of hand pixels, in order to train segmentation models online. Kumar et al. \cite{kumar2015fly} proposed an on-the-fly hand segmentation, where the user calibrated the systems by waving the hands in front of the camera. The combination of color and motion segmentation (Horn–Schunck optical flow \cite{horn1981determining}) and region growing, allowed locating the hand regions for training a GMM-based hand segmentation model. Region growing was also used by Huang et al. \cite{huang2016egocentric,huang2018egocentric}. The authors segmented the frames in super-pixels \cite{achanta2012slic} and extracted ORB descriptors \cite{rublee2011orb} from each super-pixel to find correspondences between regions of consecutive frames, which reflect the motion between two frames. Hand-related matches were distinguished from camera-related matches based on the assumption that camera-related matches play a dominant role in the video. These matches were estimated using RANSAC \cite{fischler1981random} and after being removed, those left were assumed to belong to the hands and used to locate the seed point for region growing. Zhao et al. \cite{zhao2017unsupervised,zhao2018coarse} based their approach on the typical motion pattern during actions involving the hands: preparatory phase (i.e., the hands move from the lower part of the frame to the image center) and interaction phase. During the preparatory phase they used a motion-based segmentation, computing the TV-L1 optical flow \cite{perez2013tv}. As the preparatory phase ends, the motion decreases and the appearance becomes more important. A super-pixel segmentation \cite{achanta2012slic} was then performed and a super-pixel classifier, based on the initial motion mask, was trained using color and gradient features.

Transfer learning has also been used to deal with the paucity of pixel-level annotations. The idea is to exploit the available pixel-level annotations in combination with other image segmentation techniques (e.g., super-pixels or GrabCut \cite{achanta2012slic,li2015superpixel,rother2004grabcut}) to generate new hand segmentation masks and fine-tune pre-trained networks. Zhou et al. \cite{zhou2016cascaded} trained a hand segmentation network using a large amount of bounding box annotations and a small amount of hand segmentation maps \cite{fathi2011learning}. They adopted a DeconvNet architecture \cite{noh2015learning} made up of two mirrored VGG-16 networks \cite{simonyan2014very} initialized with 1,500 pixel-level annotated frames from \cite{fathi2011learning}. Their approach iteratively selected and added good segmentation proposals to gradually refine the hand map. The hand segmentation proposals were augmented by applying super-pixel segmentation \cite{achanta2012slic} and Grabcut \cite{rother2004grabcut} to generate the hand probability map within the ground truth bounding boxes. DeconvNet was trained in an Expectation-Maximization manner: 1) keeping the network parameter fixed, they generated a set of hand masks and selected the best segmentation proposals (i.e., those with largest match with the ground truth mask); 2) they updated the network weights by using the best segmentation hypotheses. Similarly, Li et al. \cite{li2019supervised} relied on the few available pixel-level annotations to train Deeplab-VGG16 \cite{chen2017deeplab}. Their training procedure was composed of multiple steps: 1) Pre-segmentation -- the CNN, pre-trained using the available pixel-level annotations, was applied on the target images to generate pre-segmentation maps; 2) Noisy mask generation -- the pre-segmentation map was combined with a super-pixel segmentation \cite{li2015superpixel}; and 3) Model retraining -- the new masks were used as ground truth to fine tune the pre-trained Deeplab-VGG16.

\subsubsection{Depth and 3D segmentation}\label{subsubsec:seg3D}
The use of depth sensors or stereo cameras helps alleviate some of the aforementioned issues, in particular the robustness to illumination changes and lack of training data. However, the use of devices not specifically developed for wearable applications and their high power consumption \cite{betancourt2016gpu} has limited their FPV application only to research studies.

Some authors used the depth information to perform a background/foreground segmentation followed by hand/object segmentation within the foreground region by using appearance information \cite{wan2015mining,rogez2015understanding,yamazaki2017hand}. Wan et al. \cite{wan2015mining} used a time-of-flight (ToF) camera to capture the scene during hand-object interactions. They observed that the foreground (i.e., arm, hands, and manipulated objects) is usually close to the camera and well distinguishable, in terms of distance, from the background. Thus, after thresholding the histogram of depth values to isolate the foreground, hand pixels were detected by combining color and texture features (e.g., RGB thresholds and Gabor filters). The same ToF camera (Creative$^{\circledR}$ Senz3D\textsuperscript{TM}) was used by Rogez et al. \cite{rogez2015understanding}. The authors trained a multi-class classifier on synthetic depth maps of 1,500 different hand poses, in order to recognize one of these poses in the test depth images, thus producing a coarse segmentation mask. This mask was then processed in a probabilistic manner to find the binary map corresponding to the hand pixels. Color cues were also used by computing RGB-based super-pixels on the test image. Yamazaki et al. \cite{yamazaki2017hand} reconstructed the colored point cloud of the scene recorded with a Microsoft$^{\circledR}$ Kinect\textsuperscript{TM} v2. The foreground was isolated by detecting and removing large plane structures (i.e., likely belonging to the background) using RANSAC \cite{fischler1981random}. Afterwards, color segmentation was performed using a person-specific skin color model calibrated on the user’s skin. Ren et al. \cite{ren2016hand} used a stereo camera to reconstruct the depth map of the scene. Specifically, the depth map was reconstructed using the scanline-based stereo matching and the hand was segmented only using depth information.

\subsubsection{Remarks on hand segmentation}\label{subsubsec:segRemarks}
Because of the high amount of detail obtained with hand segmentation algorithms, this task is the hardest one among hand-based methods in FPV. The pixel- or super-pixel-level accuracy required for this task, combined with the intrinsic problems of egocentric vision, made this sub-area the most challenging and debated of this field of research. The effort of many researchers in finding novel and powerful approaches to obtain better results is justified by the possibility to improve not only the localization accuracy, but also to boost the performance of higher-level inference. In fact, it was demonstrated that a good hand segmentation mask can be sufficient for recognizing actions and activities involving the hands with high accuracy \cite{bambach2015lending,bambach2015viewpoint}. For this reason, pixel-level segmentation has often been used as basis of higher-inference methods.

RGB-D information can certainly improve and simplify the hand segmentation task. However, these methods are a minority with respect to the 2D counterpart, since no depth cameras have been developed for specific egocentric applications. With the recent miniaturization of depth sensors (e.g., iPhone$^{\circledR}$ X and 11) the 3D segmentation is still an area worth exploring and expanding within the next few years.

Many authors considered detection and segmentation as two steps of the same task. We preferred to split these two sub-areas given the large amount of work produced in the past few years. However, as it will be illustrated in the next section, many hand detection approaches, especially those using region-based CNNs, used the segmentation mask for generating region proposals. Perhaps, with the possibility to re-train powerful object detectors, this process has become inefficient and instead of having a “detection over segmentation”, it will be more convenient to have a “segmentation over detection”, unless the specific problem calls for a pixel-level segmentation of the entire frame. Following the great success of mask R-CNN \cite{he2017mask}, an interesting approach in this direction would be to address hand segmentation as an instance segmentation task, embedding bounding box detection and semantic segmentation of the hands in a single model.

\subsection{Hand detection and tracking}\label{subsec:handDetTrack}
Hand detection is the process of localizing the global position of the hands at frame level. This task is usually performed by fitting a bounding box around the area where the hand has been detected (see Figure \ref{fig_localization}). Hand detection allows extracting coarser information than hand segmentation, although this lower detail is counterbalanced by higher robustness to noise. If the application does not require very detailed information, this is the most popular choice as basis for hand-based higher inference. In the literature we can distinguish two main approaches: hand detection as image classification task; and hand detection as object detection task. Furthermore, hand detection generalized over time is referred to as hand tracking.

\subsubsection{Hand detection as image classification}\label{subsubsec:detImClass}
Pixel-level segmentation of hand regions, if performed on the entire image, may be prone to high occurrence of false positives \cite{betancourt2015towards,betancourt2014sequential}. In these cases, a pre-filtering step that prevents from processing frames without any hands is necessary. This approach allows determining whether an image contains hands and it is usually followed by a hand segmentation step responsible for locating the hand region \cite{betancourt2015towards,zariffa2013hand,betancourt2014sequential,zhao2017unsupervised,zhao2018coarse}.

In \cite{zariffa2013hand}, the authors back-projected the frame using a histogram obtained from a mixture of Gaussian skin model \cite{jones2002statistical}, predicting the presence of hands within the image by thresholding the back-projected values. Betancourt et al. \cite{betancourt2014sequential} proposed an approach based on HOG features and SVM classifier to predict the presence of hands at frame-level, reducing the number of false positives. However, this frame-by-frame filtering increased the risk of removing frames with barely visible hands, thus increasing the false negatives \cite{betancourt2014sequential}. To solve this issue, the authors proposed a dynamic Bayesian network (DBN) to smooth the classification results of the SVM and improve the prediction performance \cite{betancourt2015filtering}. Zhao et al. \cite{zhao2017unsupervised,zhao2018coarse} detected the presence of hands within each frame exploiting the typical interaction cycle of the hands (i.e., preparatory phase - interaction - hands out of the frame). Based on this observation, they defined an ego-saliency metric related to the probability of having hands within a frame. This metric was derived from the optical flow map calculated using \cite{margolin2013makes} and was composed of two terms: spatial cue, which gives more weight to motion within the lower part of the image; and temporal cue, which takes into account whether the motion is increasing or decreasing between adjacent frames.

\subsubsection{Hand detection as object detection}\label{subsubsec:detObjDet}
Hand detection performed within an object localization framework presents notable challenges. Given the flexibility, the continuous variation of poses, and the high number of degrees of freedom, the hand appearance is highly variable and classical object detection algorithms (e.g., Haar like features with adaboost classification) may work only in constrained situations, such as detection of hands in a specific pose \cite{wang2014finger}. For these reasons, and thanks to the availability of large annotated datasets with bounding boxes, this is the area that most benefited from the advent of deep learning.

\textbf{Region-based approaches}. Many authors proposed region-based CNNs to detect the hands, exploiting segmentation approaches summarized in Section \ref{subsec:handSeg} to generate region proposals. Bambach et al. \cite{bambach2015lending,bambach2015viewpoint} proposed a probabilistic approach for region proposal generation that combined spatial biases (e.g., reasoning on the position of the shape of the hands from training data) and appearance models (e.g., non-parametric modeling of skin color in the YUV color space). To guarantee high coverage, they generated 2,500 regions for each frame that were classified using CaffeNet \cite{jia2014caffe}. Afterwards, they obtained the hand segmentation mask within the bounding box, by applying GrabCut \cite{rother2004grabcut}. Zhu et al. \cite{zhu2016two} used a structured random forest to propose pixel-level hand probability maps. These proposals were passed to a multitask CNN to locate the hand bounding box, the shape of the hand within the bounding box, and the position of wrist and palm. In \cite{cartas2017detecting}, the authors generated region proposals by segmenting skin regions with \cite{li2013pixel} and determining if the set of segmented blobs correspond to one or two arms. This estimation was performed by thresholding the fitting error of a straight line. K-means clustering, with \textit{k} = 2 if two arms are detected, was applied to split the blobs into two separate structures. The hand proposals were selected as the top part of a rectangular bounding box fitted to the arm regions and passed to CaffeNet for the final prediction. To generate hand region proposals, Cruz et al. \cite{cruz2018hand} used a deformable part model (DPM) to make the approach robust to different gestures. DPM learns the hand shape by considering the whole structure and its parts (i.e., the fingers) using HOG features. CaffeNet \cite{jia2014caffe} was used for classifying the proposals. Faster R-CNN was used in \cite{likitlersuang2019egocentric,nguyen2018automated,nebel2018recognition}. In particular, Likitlersuang et al. \cite{likitlersuang2019egocentric} fine-tuned the network on videos from individuals with cSCI performing ADLs. False positives were removed based on the arm angle information computed by applying a Haar-like feature rotated 360 degrees around the bounding box centroid. The resulting histogram was classified with a random forest to determine whether the bounding box actually included a hand. Furthermore, they combined color and edge segmentation to re-center the bounding box, in order to promote maximum coverage of the hands while excluding parts of the forearm. 

\textbf{Regression-based approaches} were also used for detecting the hands. Mueller et al. \cite{mueller2017real} proposed a depth-based approach for hand detection, implemented using the Intel$^{\circledR}$ RealSense\textsuperscript{TM} SR300 camera. A Hand Localization Network (HALNet -- architecture derived from ResNet50 \cite{he2016deep} and trained on synthesized data) was used to regress the position of the center of the hand. The ROI was then cropped around this point based on its distance from the camera (i.e., the higher the depth, the smaller the bounding box). Recently, the \textit{You Only Look Once} (YOLO) detector \cite{redmon2016you} was applied for localizing hands in FPV \cite{visee2020effective,cruz2019my,kapidis2019egocentric}, demonstrating better trade-off between computational cost and localization accuracy than Faster R-CNN and single-shot detector (SSD) \cite{visee2020effective,cruz2019my,liu2016ssd}.

\subsubsection{Hand tracking}\label{subsubsec:detTrack} 
Hand tracking allows estimating the position of the hands across multiple frames, reconstructing their trajectories in time. Theoretically, every hand detection and segmentation approach seen above, with the exception of the binary classification algorithms of section \ref{subsubsec:detImClass}, can be used as tracker as well, by performing a frame-by-frame detection. This is the most widely used choice for tracking the hand position over time. However, some authors tried to combine the localization results with temporal models to predict the future hand positions. This strategy has several advantages, such as decreasing the computational cost by avoiding to run the hand detection every frame \cite{visee2020effective}, disambiguate overlapping hands by exploiting their previous locations \cite{lee2014hand,cai2017ego,kapidis2019egocentric}, and refining the hand location \cite{liu2016fingertip}.

Lee et al. \cite{lee2014hand} studied the child-parent social interaction from the child’s POV, by using a graphical model to localize the body parts (i.e., hands of child and parent, head of the parent). The model was composed of inter-frame links to enforce temporal smoothness of the hand positions over time, shift links to model the global shifts in the field of view caused by the camera motion, and intra-frame constraints based on the spatial configuration of the body parts. Skin color segmentation in the YUV color space was exploited to locate the hands and define intra-frame constraints on their position. This formulation forced the body parts to remain in the neighborhood of the same position between two consecutive frames, while allowing for large displacement due to global motion (i.e., caused by head movements) if this displacement is consistent with all parts. Liu et al. \cite{liu2016fingertip} demonstrated that the hand detection is more accurate in the central part of the image due to a center bias (i.e., higher number of training examples with hands in the center of the frame). To correct this bias and obtain homogeneous detection accuracy in the whole frame, they proposed an attention-based tracker (AHT). For each frame, they estimated the target location of the hand by exploiting the result at the previous frame. Then, the estimated hand region was translated to the image center, where a CNN fine-tuned on frames with centralized hands was applied. After segmenting the hand regions using \cite{li2013pixel}, Cai et al. \cite{cai2017ego} used the temporal tracking method \cite{argyros2004real} to discriminate them in case of overlap. 

Regression-based CNNs in conjunction with object tracking algorithms were used in \cite{kapidis2019egocentric,visee2020effective}. Kapidis et al. \cite{kapidis2019egocentric} fine-tuned YOLOv3 \cite{redmon2018yolov3} on multiple datasets to perform the hand detection, discriminating the right and left hand trajectories over time using the simple online real-time tracking (SORT) \cite{bewley2016simple}. For each detected bounding box, this algorithm allowed predicting its next position, also assigning it to existing tracks or to new ones. Visée et al. \cite{visee2020effective} combined hand detection and tracking to design an approach for fast and reliable hand localization in FPV. Motivated by the slow detection performance of YOLOv2 without GPU, they proposed to combine YOLOv2 with the Kernelized Correlation Filter (KCF) \cite{henriques2012exploiting} as a trade-off between speed and accuracy. The authors used the detector to automatically initialize and reset the tracker in case of failure or after a pre-defined number of frames.

\subsubsection{Remarks on hand detection and tracking}\label{subsubsec:detRemarks} 
Hand detection and segmentation are two closely related tasks that can be combined together. If hand detection is performed using a region-based approach (e.g., Faster R-CNN), hand segmentation can be seen as the pre-processing step of the localization pipeline, whereas in case of regression-based CNNs (e.g., YOLO) hand segmentation may follow the bounding box detection. The higher performance of regression-based methods with respect to region-based CNNs \cite{cruz2019my,visee2020effective} makes the latter approach more appealing in view of optimizing the hand localization pipeline. If there is no need of segmenting the hands at pixel level, the segmentation can just be skipped, whereas in problems where detailed hand silhouettes are needed, hand segmentation can be applied only within the detected the ROI, avoiding unnecessary computation.

The combination of detection and tracking algorithm may help to speed-up the localization performance with the possibility of translating these approaches into real-world application where low resource hardware is the only available option \cite{visee2020effective}. Moreover, as we will show in Section \ref{sec:interpretation}, hand tracking is an important step for the characterization and recognition of dynamic hand gestures \cite{hegde2016gestar,mohatta2017robust}.

\subsection{Hand identification}\label{subsec:handIdent} 
Hand identification is the process of disambiguating the left and right hands of the camera wearer, as well as the hands of other persons in the scene. The egocentric POV has intrinsic advantages that allow discriminating the hands by using simple spatial and geometrical constraints \cite{likitlersuang2019egocentric,lee2014hand,likitlersuang2015arm}. Usually, the user’s hands appear in the lower part of the image, with the right hand to the right of the user’s left hand, and vice versa. By contrast, other people’s hands tend to appear in the upper part of the frame \cite{lee2014hand}. The orientation of the arm regions was used in \cite{likitlersuang2019egocentric,likitlersuang2015arm} to distinguish the left from the right user’s hand. To estimate the angle, the authors rotated a Haar-like feature around the segmented hand region, making this approach robust to the presence of sleeves and different skin colors, since it did not require any calibrations \cite{likitlersuang2015arm}. To identify the hands, they split the frame into four quadrants. The quadrant with the highest sum of the Haar-like feature vector determined the hand type: “user’s right” if right lower quadrant; “user’s left” if left lower quadrant; “other hands” if upper quadrants \cite{likitlersuang2019egocentric}. The angle of the forearm/hand regions was also used by Betancourt et al. \cite{betancourt2016gpu,betancourt2017left}. The authors fitted an ellipse around the segmented region, calculating the angle between the arm and the lower frame border and the normalized distance of the ellipse center from the left border. The final left/right prediction was the result of a likelihood ratio test between two Maxwell distributions.
Although simple and effective, spatial and geometric constraints may fail in case of overlapping hands. In this case, the temporal information help disambiguate the hands \cite{cai2017ego,kapidis2019egocentric}. Cai et al. \cite{cai2017ego} were interested in studying the grasp of the right hand. After segmenting the hand regions \cite{li2013pixel}, they implemented the temporal tracking method proposed in \cite{argyros2004real} to handle the case of overlapping hands, thus tracking the right hand. Kapidis et al. \cite{kapidis2019egocentric} used the SORT tracking algorithm \cite{bewley2016simple}. This approach combines the Kalman filter to predict the future position of the hand and the Hungarian algorithm to assign the next detection to existing tracks (i.e., left/right) or new ones.

With the availability of powerful and accurate CNN-based detectors, the hand identification as separated processing step is deprecated, being incorporated within hand detection (see Section \ref{subsubsec:detObjDet}) \cite{bambach2015lending,nguyen2018automated,visee2020effective,cruz2019my}. To this end, both region-based (e.g., Faster R-CNN) and regression-based methods (e.g., YOLO and SSD) have been used. These models were trained or fine-tuned to recognize two or more classes of hands, predicting the bounding box coordinates along with its label (i.e., left, right, and other hands) \cite{visee2020effective,cruz2019my}.

\subsection{Hand pose estimation and fingertip detection}\label{subsec:handPose} 
Hand pose estimation consists in the localization of the hand parts (e.g., the hand joints) to reconstruct the articulated hand pose from the images (see Figure \ref{fig_localization}). The possibility to obtain the position of fingers, palm, and wrist, simplifies higher inference tasks such as grasp analysis and hand gesture recognition, since the dimensionality of the problem is reduced yet keeping high-detail information. An important difficulty in hand pose estimation lies in object occlusions and self-occlusions that make it hard to localize hidden joints/parts of the hand. Some authors proposed the use of depth cameras in conjunction with sensor-based techniques to train hand pose estimators more robust to self-occlusions \cite{rogez20143d,rogez2015first,mueller2017real,yamazaki2017hand,garcia2018first}. However, as discussed above, the use of RGB-D imaging techniques might not be easily translated to FPV. Thus, several attempts have also been made to estimate the hand pose using only color images \cite{liang2015egocentric,zhu2015structured,urabe2018cooking,baulig2018adapting,tekin2019h+}. In this section, we summarize the previous work distinguishing between hand pose estimation approaches with depth sensors and hand pose estimation using monocular color images. Moreover, we summarize approaches for fingertip detection, which can be seen as an intermediate step between hand detection and hand pose estimation.

\subsubsection{Hand pose estimation using depth sensors}\label{subsubsec:pose3D}
One of the advantages of using depth information for estimating the hand pose is that it is easier to synthesize depth maps that closely resemble the ones acquired by real sensors, when compared to real versus synthetic color images \cite{rogez20143d,rogez2015first}. In \cite{rogez20143d}, the authors tackled hand pose estimation as a multiclass classification problem by using a hierarchical cascade architecture. The classifier was trained on synthesized depth maps by using HOG features and tested on depth maps obtained with a ToF sensor. Instead of estimating the joint coordinates independently, they predicted the hand pose as whole, in order to make the system robust to self-occlusions. Similarly, in \cite{rogez2015first}, the authors predicted the upper arm and hand poses simultaneously, by using a multiclass linear SVM for recognizing K poses from depth data. However, instead of classifying scanning windows on the depth maps, they classified the whole egocentric work-space, defined as the 3D volume seen from the egocentric POV. Mueller et al. \cite{mueller2017real} proposed a CNN architecture (Joint Regression Net -- JORNet) to regress the 3D locations of the hand joints within the cropped colored depth maps captured with a structured light sensor (Intel$^{\circledR}$ RealSense\textsuperscript{TM} SR300). Afterwards, a kinematic skeleton was fitted to the regressed joints, in order to refine the hand pose. Yamazaki et al. \cite{yamazaki2017hand} estimated the hand pose from hand point clouds captured with the Kinect v2 sensor. The authors built a dataset by collecting pairs of hand point clouds and ground truth joint positions obtained with a motion capture system. The pose estimation was performed by aligning the test point cloud to the training examples and predicting its pose as the one that minimizes the alignment error. The sample consensus initial alignment \cite{rusu2009fast} and iterative closest point algorithms \cite{rusinkiewicz2001efficient} were used for aligning the point clouds. Garcia-Hernando et al. \cite{garcia2018first} evaluated a CNN-based hand pose estimator \cite{yuan2017bighand2} for regressing the 3D hand joints from RGB-D images recorded with the Intel$^{\circledR}$ RealSense\textsuperscript{TM} SR300 camera. The authors demonstrated that state-of-the-art hand pose estimation performance can be reached by training the algorithms on datasets that include hand-object interactions, in order to improve its robustness to self-occlusions or hand-object occlusions.

\subsubsection{Hand pose estimation from monocular color images}\label{subsubsec:poseColor}
In general, hand pose estimation from monocular color images allows locating the parts of the hands either in the form of 2D joints or semantic sub-regions (e.g., fingers, palm, etc.). This estimation is performed within previously detected ROIs, obtained by either a hand detection or segmentation algorithm. Liang et al. \cite{liang2015egocentric} used a conditional regression forest (CRF) to estimate the hand pose from hand binary masks. Specifically, they trained a set of pose estimators separately, conditioned on different distances from the camera, since the hand appearance and size can change dramatically with the distance from the camera. Thus, they synthesized a dataset in which the images were sampled at discretized intervals. The authors also proposed an intermediate step for improving the joint localization, by segmenting the binary silhouette into twelve semantic parts corresponding to different hand regions. The semantic part segmentation was performed with a random forest for pixel-level classification exploiting binary context descriptors. Similarly, Zhu et al. \cite{zhu2015structured} built a structured forest to segment the hand region into four semantic sub-regions: thumb, fingers, palm, and forearm. This semantic part segmentation was performed extending the structured regression forest framework already used for hand segmentation (as discussed in Section \ref{subsec:handSeg}) to a multiclass problem \cite{liang2015egocentric}.

Other studies adapted CNN architectures developed for human pose estimation (e.g., OpenPose \cite{wei2016convolutional,cao2017realtime}) for solving the hand pose estimation problem \cite{urabe2018cooking,baulig2018adapting} and localizing 21 hand joints. Tekin et al. \cite{tekin2019h+} used a fully convolutional network (FCN) architecture to simultaneously estimate the 3D hand and object pose from RGB images. For each frame, the FCN produced a 3D discretized grid. The 3D location of the hand joints in camera coordinate system was then estimated combining the predicted location within the 3D grid and the camera intrinsic matrix.

\subsubsection{Fingertip detection}\label{subsubsec:poseFingertip}
Fingertip detection can be seen as an intermediate step between hand detection and hand pose estimation. Unlike pose estimation, only the fingertips of one or multiple fingers are detected. These key-points alone do not allow reconstructing the articulated hand pose, but can be used as input to HCI/HRI systems \cite{song2016towards,brancati2015robust,chang2016spatio}, as will be discussed in Section \ref{sec:application}. If the objective is to estimate the joints of a single finger, the most common solution is to regress the coordinates of these points (e.g., the tip and knuckle of the index finger) from a previously detected hand ROI. This approach has been exploited in \cite{liu2016fingertip,huang2015deepfinger}. The cropped images, after being resized, were passed to a CNN to regress the location of the key-points \cite{huang2015deepfinger}. However, since the fingertip often lies at the border of the hand bounding box, the hand detection plays a significant role, and inaccurate detections greatly affect the fingertip localization result \cite{liu2016fingertip}. Wu et al. \cite{wu2017yolse} extended the fingertip detection problem to the localization of the 5 fingertips of a hand. They proposed a heatmap-based FCN that, given the detected hand area, produced a 5-channel image containing the estimated likelihood of each fingertip at each pixel location. The maximum of each channel was used to predict the position of the fingertips. 

\subsubsection{Remarks on hand pose estimation}\label{subsubsec:poseRemarks}
Among the hand localization tasks, hand pose estimation allows obtaining high-detail information with high semantic content at the same time (see Figure \ref{fig_framework}). This task, if performed correctly, can greatly simplify higher inference steps (e.g., hand gesture recognition and grasp analysis), but may be more prone to low robustness against partial hand occlusions.

Compared to other localization tasks, hand pose estimation presents a higher proportion of approaches that use depth sensors. This choice has several advantages: 1) the possibility to use motion capture methods for automatically obtaining the ground truth joint positions \cite{yamazaki2017hand,yuan2017bighand2}; 2) the availability of multiple streams (i.e., color and depth) that can be combined to refine the estimations \cite{mueller2017real,brancati2015robust}; and 3) the possibility to synthesize large datasets of realistic depth maps \cite{rogez20143d,rogez2015first}. In the past few years, human pose estimation approaches \cite{wei2016convolutional,cao2017realtime} have been successfully adapted to the egocentric POV, in order to estimate the hand and arm pose from monocular color images \cite{urabe2018cooking,baulig2018adapting}. This opens new possibilities to streamline and improve the performance of localization and hand-based higher inference tasks, such as grasp analysis. To further facilitate the adaptation of existing pose estimation approaches, large annotated datasets with hand joint information are needed. To this end, a combination of 2D and 3D information may be beneficial, in order to get accurate and extensive ground truth annotations in 3D that will allow solving the occlusion problems even when using color images alone.


\section{Interpretation}\label{sec:interpretation}
\begin{figure}
\begin{center}
\includegraphics[width=0.99\linewidth]{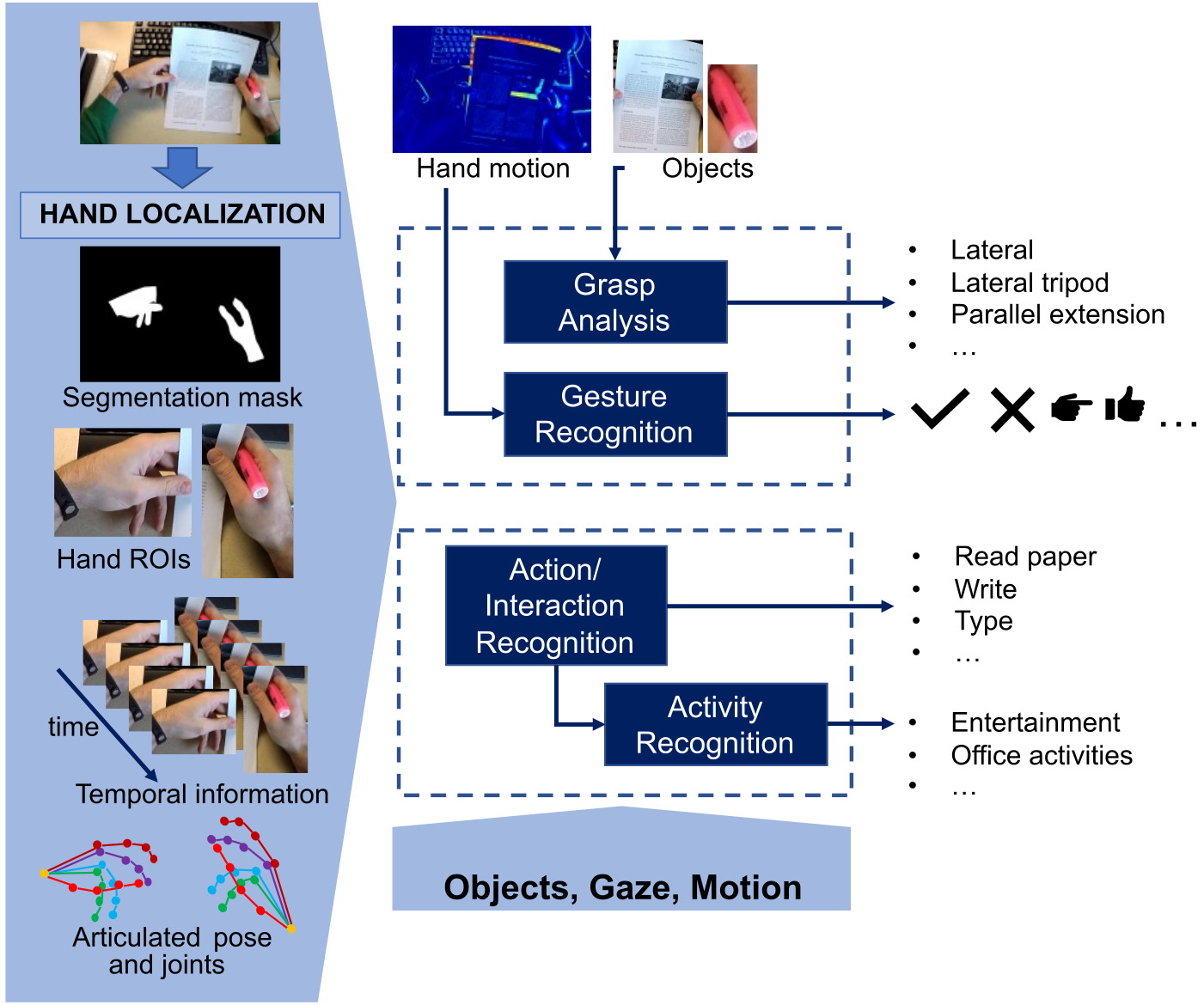}
\end{center}
\caption{Diagram of the hand interpretation areas in egocentric vision. Grasp analysis and gesture recognition focus directly on describing the hand. In action/interaction and activity recognition, the hand is instrumental in describing the user's behaviour.}
\label{fig_interpret}
\end{figure}

After the hands have been localized within the images, higher-level inference can be conducted in the ROIs. This processing is usually devoted to the interpretation of gestures and actions of the hands that, in turn, can be used as cues for hand-based applications such as HCI and HRI, as will be discussed in Section \ref{sec:application}. Based on the literature published so far, hand-based interpretation approaches in FPV can be divided into hand grasp analysis, hand gesture recognition, action/interaction recognition, and activity recognition (see Figures \ref{fig_framework} and \ref{fig_interpret}).

\subsection{Hand grasp analysis and gesture recognition}\label{subsec:graspGesture}
According to Feix et al. \cite{feix2015grasp}, “A grasp is every static hand posture with which an object can be held securely with one hand, irrespective of the hand orientation”. The recognition of the grasp types allows determining the different ways with which humans use their hands to interact with objects \cite{huang2015we}. The common grasp modes can be used to describe hand-object manipulations, reducing the complexity of the problem, since the set of possible grasps is typically smaller than the set of possible hand shapes \cite{feix2015grasp}. Moreover, the identification of the most recurrent grasp types has important applications in robotics, biomechanics, upper limb rehabilitation, and HCI. Thus, several taxonomies were proposed in the past decades \cite{feix2015grasp,feix2009comprehensive,kamakura1980patterns,cutkosky1989grasp,light1999critical,bullock2013grasp,liu2014taxonomy}. For a comprehensive comparison among these taxonomies, the reader is referred to \cite{feix2015grasp}. The analysis of hand grasps conducted via manual annotations is a lengthy and costly process. Thus, the intrinsic characteristics of egocentric vision allowed developing automated methods to study and recognize different grasp types, saving a huge amount of manual labor. Although in most cases the hand grasp analysis has been addressed in a supervised manner (i.e., grasp recognition -- Section \ref{subsubsec:graspRec}) \cite{rogez2015understanding,cai2017ego,cai2015scalable,cai2016understanding,lin2014egocentric,baydoun2017hand}, some authors proposed to tackle this problem using clustering approaches, in order to discover dominant modes of hand-object interaction and identify high-level relationships among clusters (i.e., grasp clustering and abstraction -- Section \ref{subsubsec:graspCluster}) \cite{cai2017ego,huang2015we,cai2015scalable,li2016grasp}.

Similar to grasp analysis, hand gesture recognition aims at recognizing the semantic of the hand’s posture and it is usually performed as input to HCI/HRI systems. However, two main differences exist between these two topics: 1) Grasp analysis looks at the hand posture during hand-object manipulations, whereas hand gesture recognition is usually performed on hands free of any manipulations; 2) grasp analysis aims at recognizing only static hand postures \cite{feix2015grasp}, whereas hand gesture recognition can also be generalized to dynamic gestures. According to the literature, hand gestures can be static or dynamic \cite{ren2016hand}: static hand gesture recognition (see Section \ref{subsubsec:gestureStatic}) aims at recognizing gestures that do not depend on the motion of the hands, thus relying on appearance and hand posture information only \cite{serra2013hand,ren2016hand,song2016towards,thalmann2015first,jang2016metaphoric,ji2017egocentric,zhang2018egogesture}; dynamic hand gesture recognition (see Section \ref{subsubsec:gestureDynamic}) is performed using temporal information (e.g., hand tracking), in order to capture the motion cues that allow generating specific gestures \cite{jang2016metaphoric,zhang2018egogesture,baraldi2014gesture,baraldi2015gesture,hegde2016gestar,mohatta2017robust}.

\subsubsection{Hand grasp recognition}\label{subsubsec:graspRec}
Supervised approaches for grasp recognition are based on the extraction of features from previously segmented hand regions \cite{li2013pixel} and their multiclass classification following one of the taxonomies proposed in the literature \cite{feix2015grasp}. 

Cai et al. \cite{cai2015scalable} used HOG features to represent the shape of the hand and a combination of HOG and SIFT to capture the object context during the manipulation. These features were classified with a multi-class SVM using a subset of grasp types from Feix’s taxonomy \cite{feix2015grasp,feix2009comprehensive}. The authors extended their approach in \cite{cai2017ego,cai2016understanding} by introducing CNN-based features extracted from the middle layers of \cite{krizhevsky2012imagenet} and features derived from the dense hand trajectory (DHT) \cite{wang2011action} such as the displacement, gradient histograms, histogram of optical flow, and motion boundary histograms. The superior performance of CNN- and DHT-based features and their robustness across different tasks and users \cite{cai2017ego} suggested that high-level feature representation and motion and appearance information in the space-time volume may be important cues for discriminating different hand configurations. In \cite{baydoun2017hand}, the authors used a graph-based approach to discriminate 8 grasp types. Specifically, the binary hand mask was used to produce a graph structure of the hand with an instantaneous topological map neural network. The eigenvalues of the graph’s Laplacians were used as features to represent the hand configurations, which were recognized using an SVM. 

The use of depth sensors was explored by Rogez et al. \cite{rogez2015understanding}. The authors recognized 71 grasp types \cite{liu2014taxonomy} using RGB-D data, by training a multi-class SVM with deep-learned features \cite{simonyan2014very} extracted from both real and synthetic data. Moreover, the grasp recognition results were refined by returning the closest synthetic training example, namely the one that minimized the distance with the depth of the detected hand region.

\subsubsection{Hand grasp clustering and abstraction}\label{subsubsec:graspCluster}
The first attempt to discover hand grasps in FPV was \cite{huang2015we}. HOG features were extracted from previously segmented hand regions and grouped by means of a two-stage clustering approach. First, a set of candidate cluster centers was generated through the fast determinantal point process (DPP) algorithm \cite{kulesza2012determinantal}. This step allowed generating a wide diversity of clusters to cover many possible hand configurations. Secondly, each segmented region was assigned to the nearest cluster center. The use of the DPP algorithm was proven to outperform other clustering approaches such as k-means and to be more appropriate in situations, like grasp analysis, where certain clusters are more recurrent than other ones. A hierarchical structure of the grasp types was learned using the same DPP-based clustering approach \cite{huang2015we}. A hierarchical clustering approach was also used in \cite{li2016grasp} to find the relationships between different hand configurations based on a similarity measure between pairs of grasp types. Similarly, in \cite{cai2017ego,cai2015scalable}, the authors used a correlation index to measure the visual similarity between grasp types: grasp types with high correlation were clustered at the lower nodes, whereas low-correlated types were clustered higher in the hierarchy. The above approaches \cite{cai2017ego,huang2015we,cai2015scalable,li2016grasp} were used to build tree-like structures of the grasp types. These structures can be exploited to define new taxonomies depending on the trade-off between detail and robustness of grasp classification, as well as to discover new grasp types not included in previous categorizations \cite{cutkosky1989grasp}.

\subsubsection{Static hand gesture recognition}\label{subsubsec:gestureStatic}
The recognition of static hand gestures is usually performed in a supervised manner, similarly to the approaches presented in Section \ref{subsubsec:graspRec} for hand grasp recognition. A common strategy is to exploit features extracted from previously segmented hand regions, classifying them into multiple gestures often using SVM classifiers \cite{serra2013hand,ren2016hand}. 

Serra et al. \cite{serra2013hand} classified the binary segmentation masks into multiple hand configurations by using an ensemble of exemplar-SVMs \cite{malisiewicz2011ensemble}. This approach was proven to be robust in case of unbalanced classes, like hand gesture recognition applications where most of the frames contain negative examples. Contour features were used in \cite{ren2016hand} to recognize 14 gestures. The authors described the silhouette of the hand shape using time curvature analysis and fed an SVM classifier with the extracted features. The use of CNNs has also been investigated for the recognition of static hand gestures \cite{baydoun2017hand,ji2017egocentric}. Ji et al. \cite{ji2017egocentric} used a hybrid CNN-SVM approach, where the CNN was implemented as feature extractor and the SVM as gesture recognizer. In \cite{song2016towards}, the authors proposed a CNN architecture to directly classify the binary hand masks into multiple gestures.

Depth information was used in \cite{thalmann2015first,jang2016metaphoric,zhang2018egogesture}. In \cite{thalmann2015first} the authors used depth context descriptors and random forest classification, whereas Jang et al. \cite{jang2016metaphoric} implemented static-dynamic voxel features to capture the amount of point clouds within a voxel, in order to describe the static posture of the hands and fingers. Moreover, depth-based gesture recognition was demonstrated to be more discriminative than color-based recognition \cite{zhang2018egogesture}. However, in addition to the drawbacks of wearable depth sensors already discussed in the previous sections, the performance were significantly lower in outdoor environments due to the deterioration of the depth map \cite{zhang2018egogesture}.

\subsubsection{Dynamic hand gesture recognition}\label{subsubsec:gestureDynamic}
One of the most common choices for dynamic hand gesture recognition is to use optical flow descriptors from the segmented hand regions, in order to recognize the motion patterns of the gestures to be classified \cite{baraldi2014gesture,baraldi2015gesture,hegde2016gestar,mohatta2017robust}. 

Baraldi et al. \cite{baraldi2014gesture,baraldi2015gesture} developed an egocentric hand gesture classification system able to recognize the user’s interactions with artworks in a museum. After removing camera motion, they computed and tracked the feature points at different spatial scales within the hand ROI and extracted multiple descriptors from the obtained spatio-temporal volume (e.g., HOG, HOF, and MBH). Linear SVM was used for recognizing multiple gestures from the above descriptors, using Bag of Words (BoW) and power normalization to avoid sparsity of the features. In \cite{hegde2016gestar,mohatta2017robust} the flow vectors were calculated over the entire duration of a gesture and, based on the resultant direction of the flow vectors, different swipe movements (e.g., left, right, up, and down) were classified using fixed thresholds on the movement orientation.

Other approaches recognized dynamic gestures as generalization of the static gesture recognition problem \cite{jang2016metaphoric,zhang2018egogesture}. In \cite{jang2016metaphoric}, the authors proposed a hierarchical approach for estimating hand gestures using a static-dynamic forest to produce hierarchical predictions on the hand gesture type. Static gesture recognition was performed at the top level of the hierarchy, in order to select a virtual object corresponding to the detected hand configuration (e.g., holding a stylus pen). Afterwards, the recognition of dynamic gestures, conditioned to the previously detected static gesture, was performed (e.g., pressing or releasing the button on the pen). Zhang et al. \cite{zhang2018egogesture} compared engineered features and deep learning approaches (e.g., 2DCNN, 3DCNN, and recurrent models), demonstrating that 3DCNN are more suitable for dynamic hand gesture recognition and the combination of color and depth information can produce better results than the two image modalities alone. 

\subsubsection{Remarks on hand grasp analysis and gesture recognition}\label{subsubsec:graspGestureRemarks}
Many similarities can be found between grasp recognition and hand gesture recognition. As mentioned above, the main difference is the context in which the two problems are addressed. Grasp recognition is performed during hand-object manipulations, whereas hand gesture recognition is performed without the manipulation of physical objects. This difference links these two sub-areas to some of the higher levels and FPV applications. In fact, hand gesture recognition approaches have mainly been used for AR/VR applications \cite{hegde2016gestar,thalmann2015first,jang2016metaphoric,mohatta2017robust}, whereas grasp analysis can be exploited for action/interaction recognition and activity recognition \cite{cai2016understanding,coskun2019domain}. In particular, the contextual relationship between grasp types and object attributes, such as rigidity and shape, has motivated authors \cite{cai2015scalable,cai2016understanding} to exploit object cues for improving the grasp recognition performance.

Hand grasp analysis and gesture recognition are the only interpretation sub-areas where the analysis of the hands is still the main target of the approaches. In fact, higher in the semantic content dimension, sub-areas like action recognition, interaction detection, and activity recognition may use the hand information in combination with other cues (e.g., object recognition) to perform higher level inference. It should be noted though, that not all the higher-level interpretation approaches utilized hand-based processing in FPV. Thus, in the following sections, we will discuss only those methods that explicitly used the hand information for predicting actions and activities, omitting other papers and referring the authors to other surveys or research articles.

\subsection{Action/interaction and activity recognition}\label{subsec:actionActivity}
According to Tekin et al. \cite{tekin2019h+}, an action is a verb (e.g., “cut”), whereas an interaction is a verb-noun pair (e.g., “cut the bread”). Both definitions refer to short-term events that usually last a few seconds \cite{singh2016first}. By contrast, activities are longer temporal events (i.e., minutes or hours) with higher semantic content, typically composed of temporally-consistent actions and interactions \cite{nguyen2016recognition} (see Figure \ref{fig_interpret}).

In this section, we summarize FPV approaches that relied on hand information to recognize actions, interactions, and activities from sequences of frames. Regarding the actions and interactions, two main types of approaches can be found in literature: those that used hands as the only cue for the prediction (see Section \ref{subsubsec:actionHand}) and approaches that used a combination of object and hand cues (see Section \ref{subsubsec:actionHandObj}). Although the second type of approaches might seem more suitable for interaction recognition (i.e., verb + noun prediction), some authors used them for predicting action verbs, exploiting the object information to prune the space of possible actions (i.e., removing unlikely verbs for a given object) \cite{fathi2011understanding}. Likewise, other authors tried to use only hand cues to recognize interactions \cite{tang2018multi}, in order to produce robust predictions without relying on object features or object recognition algorithms. Either way, the boundary between action and interaction recognition is not well defined and often depends on the nature of the dataset on which a particular approach has been tested.

\subsubsection{Action/interaction recognition using hand cues}\label{subsubsec:actionHand}
These approaches inferred the camera wearer’s actions exploiting the information provided by hand localization methods. The hypothesis is that actions and interactions can be recognized using only hand cues, for instance features related to the posture and motion of the hands. Existing studies can be divided into feature-based approaches \cite{kumar2015fly,ishihara2015recognizing,cai2018desktop} and deep learning-based approaches \cite{singh2016first,urabe2018cooking,tang2018multi}.

Feature-based approaches combined motion and shape features of the hands to represent two complementary aspects of the action: movements of hand’s parts and grasp types. This representation allowed discriminating actions with similar hand motion, but different hand posture. Typical choices of motion features were dense trajectories \cite{wang2011action}, whereas the hand shape was usually represented with HOG \cite{ishihara2015recognizing} or shape descriptors on the segmented hand mask \cite{cai2018desktop}. All these features were then combined and used to recognize actions/interactions via SVM classifiers. Ishihara et al. \cite{ishihara2015recognizing} used dense local motion features to track keypoints from which HOG, MBH, and HOF were extracted \cite{wang2011action}. Global hand shape was represented using HOG features within the segmented hand region. The authors used Fisher vectors and principal component analysis to encode features extracted from time windows of fixed duration, followed by multiclass linear SVM for the recognition. Dense trajectory features were also used by Kumar et al. \cite{kumar2015fly}. The authors proposed a feature sampling scheme that preserved dense trajectories closer to the hand centroid while removing trajectories from the background, which are likely caused by head motion. BoW representation was used and the recognition was performed using SVM with $\chi^{2}$ kernel. Cai et al. \cite{cai2018desktop} combined hand shape, hand position, and hand motion features for recognizing user’s desktop actions (e.g., browse, note, read, type, and write). Histograms of the hand shape computed on the hand mask were used as shape features. Hand position was represented by the point within the hand region where a manipulation is most likely to happen (e.g., left tip of the right hand region). Motion descriptors relied on the computation of the large displacement optical flow (LDOF) \cite{brox2010large} between two consecutive frames. Spatio-temporal distribution of hand motion (i.e., discrete Fourier transform coefficients on the average LDOF extracted from hand sub-regions over consecutive frames) was demonstrated to outperform temporal and spatial distributions alone, suggesting that spatial and temporal information should be considered together when recognizing hand’s actions.

The combination of temporal and spatial information was also exploited in deep-learning approaches. This strategy was usually implemented by means of multi-stream architectures. Singh et al. \cite{singh2016first} proposed a CNN-based approach to recognize camera wearer’s actions using the following inputs: pixel-level hand segmentation mask; head motion -- as frame-to-frame homography using RANSAC on optical flow correspondences excluding the hand regions; and saliency map -- as the flow map obtained after applying the homography. This information was passed to a 2-stream architecture composed of a 2DCNN and a 3DCNN. The deep-learned features from both streams were combined and actions were predicted using SVM. Urabe et al. \cite{urabe2018cooking} used the region around the hands to recognize cooking actions. Appearance and motion maps were obtained using the segmented hand mask passed to 2DCNN and 3DCNN, respectively. Afterwards, class-score fusion was performed by multiplying the output of both streams. The authors demonstrated that a multi-stream approach yielded better results than the two streams alone. Tang et al. \cite{tang2018multi} used the hand information as auxiliary stream within an end-to-end multi-stream deep neural network (MDNN) that used RGB, optical flow and depth maps as input. The hand stream was composed of a CNN with the hand mask as input. Its output was combined to the MDNN via weighted fusion, in order to predict the action label. The addition of the hand stream improved the recognition performance.

\subsubsection{Action/interaction recognition combining hand and object cues}\label{subsubsec:actionHandObj}
Many authors demonstrated that the combination of object and hand cues can improve the recognition performance \cite{fathi2011understanding,li2015delving,wan2015mining,cai2016understanding,kapidis2019egocentric}. This is quite intuitive, since during an interaction the grasp type and hand movements strictly depend on the characteristics of the object that is being manipulated (e.g., dimension, shapes, functionality) \cite{cai2016understanding}. Thus, grasp type or hand pose/shape along with object cues can be used to recognize the actions and interactions \cite{tekin2019h+,fathi2011understanding,likitlersuang2019egocentric,garcia2018first,cai2016understanding,coskun2019domain}.

In \cite{cai2016understanding}, the authors predicted the attributes of the manipulated object (i.e., object shape and rigidity) and the type of grasp to recognize hand’s actions. They proposed a hierarchical 2-stage approach where the lower layer -- visual recognition -- classified the grasp type and the object attributes and pass this information to the upper layer -- action modeling -- responsible for the action classification via linear SVM. Coskun et al. \cite{coskun2019domain} implemented a recurrent neural network (RNN) to exploit the temporal dependencies of consecutive frames using a set of deep-learned features related to grasp, optical flow, object-object, and hand-object interactions, as well as the trajectories of the hands over the past few frames. Other authors \cite{tekin2019h+,fathi2011understanding,likitlersuang2019egocentric,garcia2018first}, used hand cues with lower semantic content than hand grasp, such as shape and pose. Fathi et al. \cite{fathi2011understanding} extracted a set of object and hand descriptors (e.g., object and hand labels, optical flow, location, shape, and size) at super-pixel level and performed a 2-stage interaction recognition. First, they recognized actions using Adaboost; second, they refined the object recognition in a probabilistic manner by exploiting the predicted verb label and object classification scores. Likitlersuang et al. \cite{likitlersuang2019egocentric} detected the presence of interactions between the camera wearer’s hands and manipulated objects. This was accomplished by combining the hand shape, represented with HOG descriptors, with color and motion descriptors (e.g., color histogram and optical flow) for the hand, the background, and the object (i.e., regions around the hands). Random forest was used for classification. The articulated hand pose was used in \cite{tekin2019h+,garcia2018first}. Garcia-Hernando et al. \cite{garcia2018first} passed the hand and object key-points to an LSTM that predicted the interactions over the video frames. This approach was extended in \cite{tekin2019h+}, where hand-object interactions were first modeled using a multi-layer perceptron and then used as input to the LSTM.

Other approaches, instead of explicitly using the hand information for predicting actions and interactions, exploited the results of hand localization algorithms to guide the feature extraction within a neighborhood of the manipulation region \cite{li2015delving,ma2016going}. This strategy was motivated by the fact that the most important cues (i.e., motion, object, etc.) during an action are likely to be found in proximity of the hands and manipulated object. Li et al. \cite{li2015delving} used a combination of local descriptors for motion and object cues in conjunction with a set of egocentric cues. The former, were extracted from the dense trajectories to represent the motion of the action (i.e., shape of the trajectories, MBH, HoF) and the object appearance (e.g., HOG, LAB color histogram, and LBP along the trajectories). The latter were used to approximate the gaze information, by combining camera motion removal and hand segmentation, in order to focus the attention on the area where the manipulation is happening. Ma et al. \cite{ma2016going} used a multi-stream deep learning approach composed of an appearance stream to recognize the object and a motion stream to predict the action verb. The object recognition network predicted the object label by using as input the hand mask and object ROI, whereas the action recognition network used the optical flow map to infer the verb. A fusion layer combined verb and object labels and predicted the interactions. Zhou et al. \cite{zhou2016cascaded} used the hand segmentation mask, object features extracted from middle layers of AlexNet \cite{krizhevsky2012imagenet}, and optical flow to localize and recognize the active object using VGG-16 \cite{simonyan2014very}. Afterwards, object features were represented in a temporal pyramid manner and combined with motion characteristics extracted from improved dense trajectories, in order to recognize interactions using non-linear SVM. 
Although the above approaches might differ for the type of features and algorithm used to predict actions and interactions, most of them demonstrated that the combination of object and hand cues can provide better recognition performance than single modality recognition \cite{li2015delving,wan2015mining}.

\subsubsection{Activity recognition}\label{subsubsec:activity}
As we climb the semantic content dimension in the proposed framework, the strong dependency on hand cues fades away. Other information comes into play and can be used in conjunction with the hands to predict the activities. This diversification becomes clear when we look at the review published by Nguyen et al. \cite{nguyen2016recognition}, which categorized egocentric activity recognition as: 1) combination of actions; 2) combination of active objects; 3) combination of active objects and locations; 4) combination of active objects and hand movements; and 5) combination of other information (e.g., gaze, motion, etc.). The description of all these approaches goes beyond the scope of this work, since we are interested in characterizing how hands can be used in activity recognition methods. For a more comprehensive description of activity recognition in FPV, the reader is referred to \cite{nguyen2016recognition}.
The boundary between the recognition of short and long temporal events (i.e., actions/interactions and activities, respectively) is not always well defined and, similar to action/interaction recognition, it may depend on the dataset used for training and testing a particular approach. In fact, some of the methods described in the previous subsections were also tested within an activity recognition framework \cite{wan2015mining,ma2016going}.
Generally, we can identify two types of approaches: activity recognition based on actions and interactions \cite{fathi2011understanding,nebel2018recognition} and approaches that used hand localization results to directly prectict the activities \cite{bambach2015lending,urooj2018analysis,bambach2015viewpoint}. 

Approaches that relied on actions and interactions learned a classifier for recognizing the activities using the detected actions or hand-object interactions as features for the classification. This can be performed by using the histogram of action frequency in the video sequence and its classification using adaboost \cite{fathi2011understanding}. Nguyen et al. \cite{nebel2018recognition} used Bag of Visual Words representation to model the interactions between hands and objects, since these cues play a key role in the recognition of activities. Dynamic time warping was then used to compare a new sequence of features with the key training features.

Other authors \cite{bambach2015lending,urooj2018analysis,bambach2015viewpoint} investigated how good the hand segmentation map is in predicting a small set of social activities, such as 4 interactions between two individuals. The authors used a CNN-based approach using the binary hand segmentation maps as input. The prediction was performed on a frame-by-frame basis and using temporal integration implemented through a voting strategy among consecutive frames, with the latter approach providing better results (up to 73\% of recognition accuracy) \cite{bambach2015lending}. This result confirms what was already shown for actions and interactions, namely the temporal information becomes essential when performing higher-level inference, especially when modeling relatively long term events like activities. However, this approach was tested only in case of a small sample of social activities. To the best of our knowledge, no experiments using hand cues only were conducted for predicting other types of activities, such as ADLs.

\subsubsection{Remarks on action/interaction and activity recognition}\label{subsubsec:actionActivityRemarks}
Many authors demonstrated that action/interaction recognition performance can be improved by combining different cues, such as hand, object, and motion information. This was proven regardless of the actual method. In fact, both feature-based and deep-learning based methods implemented this strategy by combining multiple features or using multi-stream deep-learning approaches. Recently, multi-task learning approaches have also been proposed for solving the action recognition problem \cite{kapidis2019multitask}, demonstrating that predicting hand coordinates as an auxiliary task leads to an improvement in verb recognition performance with respect to the single-task approach. In the near future, it will be interesting to compare multi-task and multi-stream architectures to understand whether the joint prediction of action labels and hand positions can actually provide state-of-the-art performance in one or both tasks.

Another important aspect on which one should focus when developing novel approaches for action/interaction recognition is the temporal information. This was exploited by using 3DCNN and RNNs or, in case of feature-based approaches, by encoding it in the motion information. The same conclusion can be drawn for activity recognition where, considering the longer duration of the events, the temporal information becomes even more important \cite{bambach2015lending}.

Sometimes the literature is not consistent on the choice of the taxonomy to describe these sub-areas. Some of the approaches summarized above, even though not explicitly referred as action/interaction recognition, actually recognized short actions or interactions. We preferred to be consistent with the definition proposed by Tekin et al. \cite{tekin2019h+}, as we believe that a consistent taxonomy may help authors comparing different approaches and unify their efforts towards solving a specific problem. Moreover, the term “action” has often been used interchangeably with “activity”, which indicates a longer event with higher semantic content. The actions and interactions can rather be seen as the building blocks of the activities. This allowed some authors to exploit this mutual dependency, in order to infer activities in a hierarchical manner, using the methods described above \cite{fathi2011understanding,nebel2018recognition}.

The number of egocentric activity recognition approaches based on hand information is lower than the number of action and interaction recognition approaches. This difference is due to the fact that higher in the semantic content, authors have a wider choice of cues and features for recognizing a temporal event. In particular, over the past few years, more and more end-to-end approaches for activity recognition have been proposed, similarly to video recognition \cite{rodriguez2019video}.


\section{Application}\label{sec:application}
The hand-based approaches summarized so far can be implemented to design real-world FPV applications. Most of these applications relied on HCI and HRI and included AR and VR systems \cite{ha2014wearhand,brancati2015robust}, robot control and learning \cite{aksoy2015semantic,song2016towards}, as well as healthcare applications \cite{likitlersuang2019egocentric,nguyen2018automated}.

\subsection{AR and VR applications}
Given the recent success of VR headsets and the surge of AR applications, many hand-based methods in FPV were used for AR and VR systems to design natural user interfaces. Most of these applications relied on hand localization -- in particular, hand detection, segmentation, and pose-estimation -- and gesture recognition algorithms, for example to manipulate virtual objects in an AR or VR scenario \cite{ha2014wearhand,jang20153d,jang2016metaphoric,thalmann2015first}. Depth sensors were usually implemented to capture the scene, whereas head-worn displays allowed projecting the virtual object in the AR/VR scenario. The recognition of specific hand gestures allowed providing inputs and commands to the system, in order to produce a specific action (e.g., the selection of a virtual object by recognizing the clicking gesture \cite{jang20153d}). The use of depth sensors has usually been preferred to RGB cameras since the localization of hands and objects can be more robust to illumination changes. Some authors even implemented multiple depth sensors \cite{ha2014wearhand}: one specific for short distances (i.e., up to 1 m) -- to capture more accurate hand information -- and a long-range depth camera to reproduce the correct proportions between the physical and virtual environment \cite{ha2014wearhand}. To improve the hand localization robustness, other systems combined multiple hand localization approaches, such as hand pose estimation in conjunction with fingertip detection \cite{jang20153d}. This approach can be helpful when the objective is to localize the fingertips in situations with frequent self-occlusions. Other AR/VR applications relied on dynamic hand gestures (e.g. swipe movements) recorded with a smartphone camera and frugal VR devices (e.g., Google Cardboard), in order to enable interactions in the virtual environment \cite{hegde2016gestar,mohatta2017robust}. AR/VR applications were also implemented for tele-support and coexistence reality \cite{yu2015hand,gupta2017hand}, in order to allow multiple users to collaborate together remotely. Specific fields of applications were remote co-surgery \cite{yu2015hand} and expert's tele-assistance and support \cite{gupta2017hand}.

Within the AR context, the use of hand-based information was also exploited for recognizing hand-written characters \cite{chang2016spatio,huang2016pointing}. This application was performed in four steps: 1) hand localization through hand detection and tracking; 2) hand gesture recognition -- to recognize a specific hand posture that triggers the writing module; 3) fingertip detection -- to identify the point to track, whose trajectory defines the characters; and 4) character recognition, based on the trajectories of the detected fingertip \cite{chang2016spatio,huang2016pointing}.

Hand localization and gesture recognition approaches were also used for cultural heritage applications to develop systems for immersive museum and augmented touristic experiences \cite{serra2013hand, baraldi2014gesture, baraldi2014gesture, brancati2015robust}. Users can experience an entertaining way of accessing the museum knowledge, for example by taking pictures and providing feedback to the artworks with simple hand gestures \cite{baraldi2015gesture}. Other authors \cite{brancati2015robust} proposed a smart glasses-based system that allowed users to access touristic information while visiting a city by using pointing gestures of the hand.

\subsection{Robot control and learning}
In the HRI field, FPV hand-based approaches have mainly been used for two purposes: robot learning and robot control. Approaches for robot learning recognized movements and/or actions performed by the user's hands, in order to train a robot performing the same set of movements autonomously \cite{aksoy2015semantic,lee2017learning}. Aksoy et al. \cite{aksoy2015semantic}, decomposed each manipulation into shorter chunks and encoded each manipulation into a semantic event chain (SEC), which encodes the spatial relationship between objects and hands in the scene. Each temporal transition in the SEC (e.g., change of state in the scene configuration) was considered as movement primitive for the robot imitation. In \cite{lee2017learning}, the robot used the tracked hand locations of a human to learn the hand's future position and predict trajectories of the hands when a particular action has to be executed. By contrast, robot control approaches mainly relied on hand gesture recognition to give specific real-time commands to the robots \cite{song2016towards,ji2017egocentric}. The hand gestures are seen as means of communication between the human and the robot and can encode specific commands such as the action to be performed by a robot arm \cite{song2016towards} or the direction to be taken by a reconnaissance robot \cite{ji2017egocentric}.

\subsection{Remote healthcare monitoring}\label{subsec:appHealth}
Egocentric vision has demonstrated the potential to have an important impact in healthcare. The possibility to automatically analyze the articulated hand pose and recognize actions and ADLs have made these methods appealing for the remote assessment of upper limb functions \cite{zariffa2013hand,likitlersuang2016interaction,likitlersuang2017views,likitlersuang2019egocentric,visee2020effective} and AAL systems \cite{karaman2014hierarchical,nguyen2016recognition,nguyen2018automated}.
The assessment of upper limb function is an important phase in the rehabilitation after stroke or cSCI that allows clinicians to plan the optimal treatment strategy for each patient. However, geographical distances between patients and hospitals create barriers towards obtaining optimal assessments and rehabilitation outcomes. Egocentric vision has inspired researchers to develop video-based approaches for automatically studying hand functions at home \cite{zariffa2013hand,likitlersuang2016interaction,likitlersuang2019egocentric,visee2020effective}. Studies have been conducted in individuals with cSCI, tackling the problem of hand function assessment from two perspectives: localization \cite{zariffa2013hand,likitlersuang2015arm,visee2020effective} and interpretation \cite{likitlersuang2016interaction,likitlersuang2019egocentric}. Fine-tuning object detection algorithms to localize and recognize hands in people with SCI allowed developing hand localization approaches robust to impaired hand poses and uncontrolled situations \cite{visee2020effective}. Moreover, strategies for improving the computational performance of hand detection algorithms have been adopted (e.g., combining hand detection and tracking), making this application suitable for the use at home. The automatic detection of hand-object manipulations allowed extracting novel measures reflective of the hand usage at home, such as number of interactions per hour, the duration of interactions, and the percentage of interaction over time \cite{likitlersuang2019egocentric}. These measures, once validated against clinical scores, will help clinicians to better understand how individuals with cSCI and stroke use their hands at home while performing ADLs.

Another healthcare application is the development of AAL systems. The increasing ageing population is posing serious social and financial challenges in many countries. These challenges have stimulated the interest in developing technological solutions to help and support older adults with and without cognitive impairments during their daily life \cite{rashidi2012survey}. Some of these applications used egocentric vision to provide help and support to older adults during ADLs at home \cite{karaman2014hierarchical,nguyen2016recognition,nguyen2018automated}. Egocentric vision AAL builds upon the action and activity recognition approaches illustrated in Section \ref{subsec:actionActivity}. In particular, approaches have been proposed to automatically recognize how older adults perform ADLs at home, for example to detect early signs of dementia \cite{karaman2014hierarchical} or to support people in conducting the activities \cite{nguyen2018automated}.

In these specific applications, the use of egocentric vision presents important advantages with respect to other solutions (e.g., sensor-based and third person vision):
\begin{itemize}[noitemsep, wide=0pt, leftmargin=\dimexpr\labelwidth + 2\labelsep\relax]
    \item FPV can provide high quality videos on how people manipulate objects. This is important when the aim is the recognition of hand-object manipulations and ADLs, since hand occlusions tend to be minimized.
    \item Egocentric vision provides more details of hand-object interactions than sensor-based technology, by capturing information about both the hand and the object being manipulated. Other sensor-based solutions such as sensor gloves, although providing highly accurate hand information, may limit movements and sensation, which are already reduced in individuals with upper limb impairment \cite{likitlersuang2019egocentric,likitlersuang2017views}.
\end{itemize}


\section{FPV Datasets with hand annotation}\label{sec:datasets}

\begin{table*}[]
\begin{center}
\begin{tabular}{|ccccccccccc|}
\hline
Dataset & Year &  Mode & Device & Location & Frames & Videos & Duration & Subjects & \begin{tabular}[c]{@{}c@{}}Resolution\\ (pixels)\end{tabular} & Annotation \\ \hline\hline
GTEA \cite{fathi2011learning} & 2011 & C & GoPro & H & $\sim$31k & 28 & 34 min & 4 & 1280$\times$720 & \begin{tabular}[c]{@{}c@{}} act\\ msk\end{tabular} \\ \hline
ADL \cite{pirsiavash2012detecting} & 2012 & C & GoPro & H & \textgreater{}1M & 20 & $\sim$10 h & 20 & 1280$\times$960 & \begin{tabular}[c]{@{}c@{}}act\\ obj\end{tabular} \\ \hline
EDSH \cite{li2013pixel} & 2013 & C & - & H & $\sim$20k & 3 & $\sim$10 min & - & 1280$\times$720 & msk\\ \hline
\begin{tabular}[c]{@{}c@{}}Interactive\\ Museum \cite{baraldi2014gesture}\end{tabular} & 2014 & C & - & H & - & 700 & - & 5 & 800$\times$450 & \begin{tabular}[c]{@{}c@{}}gst\\ msk\end{tabular} \\ \hline
EgoHands \cite{bambach2015lending} & 2015 & C & Google Glass & H & $\sim$130k & 48 & 72 min & 8 & 1280$\times$720 & msk\\ \hline
Maramotti \cite{baraldi2015gesture} & 2015 & C & - & H & - & 700 & - & 5 & 800$\times$450 & \begin{tabular}[c]{@{}c@{}}gst\\ msk\end{tabular} \\ \hline
\begin{tabular}[c]{@{}c@{}}UNIGE\\ Hands \cite{betancourt2015dynamic}\end{tabular} & 2015 & C & \begin{tabular}[c]{@{}c@{}}GoPro\\ Hero3+\end{tabular} & H & $\sim$150k & - & 98 min & - & 1280$\times$720 & det\\ \hline
GUN-71 \cite{rogez2015understanding} & 2015 & CD & \begin{tabular}[c]{@{}c@{}}Creative\\ Senz3D\end{tabular} & C & \begin{tabular}[c]{@{}c@{}}$\sim$12k\\ (annotated)\end{tabular} & - & - & 8 & - & grs\\ \hline
\begin{tabular}[c]{@{}c@{}}RGBD\\ Egocentric\\ Action \cite{wan2015mining}\end{tabular} & 2015 & CD & \begin{tabular}[c]{@{}c@{}}Creative\\ Senz3D\end{tabular} & H & - & - & - & 20 & \begin{tabular}[c]{@{}c@{}}C:640$\times$480\\ D:320$\times$240\end{tabular} & act\\ \hline
\begin{tabular}[c]{@{}c@{}}Fingerwriting\\ in mid-air \cite{chang2016spatio}\end{tabular} & 2016 & CD & \begin{tabular}[c]{@{}c@{}}Creative\\ Senz3D\end{tabular} & H & $\sim$8k & - & - & - & - & \begin{tabular}[c]{@{}c@{}}ftp\\ gst\end{tabular} \\ \hline
Ego-Finger \cite{huang2016pointing} & 2016 & C & - & H & $\sim$93k & 24 & - & 24 & 640$\times$480 & \begin{tabular}[c]{@{}c@{}}det\\ ftp\end{tabular} \\ \hline
\begin{tabular}[c]{@{}c@{}}ANS\\ able-\\ bodied \cite{likitlersuang2016interaction}\end{tabular} & 2016 & C & Looxcie 2 & - & - & - & 44 min & 4 & 640$\times$480 & int\\ \hline
UT Grasp \cite{cai2017ego} & 2017 & C & \begin{tabular}[c]{@{}c@{}}GoPro\\ Hero2\end{tabular} & H & - & 50 & $\sim$4 h & 5 & 960$\times$540 & grs\\ \hline
GestureAR \cite{mohatta2017robust} & 2017 & C & \begin{tabular}[c]{@{}c@{}}Nexus 6 and\\ Moto G3\end{tabular} & H & - & 100 & - & 8 & 1280$\times$720 & gst\\ \hline
EgoGesture \cite{wu2017yolse} & 2017 & C & - & - & $\sim$59k & - & - & - & - & \begin{tabular}[c]{@{}c@{}}det\\ ftp\\ gst\end{tabular} \\ \hline
\begin{tabular}[c]{@{}c@{}}Egocentric\\ hand-action \cite{xu2017hand}\end{tabular} & 2017 & D & \begin{tabular}[c]{@{}c@{}}Softkinetic\\ DS325\end{tabular} & H & $\sim$154k & 300 & - & 26 & 320$\times$240 & gst\\ \hline
BigHand2.2M \cite{yuan2017bighand2} & 2017 & D & \begin{tabular}[c]{@{}c@{}}Intel\\ RealSense\\ SR300\end{tabular} & - & $\sim$290k & - & - & - & 640$\times$480 & pos\\ \hline
\begin{tabular}[c]{@{}c@{}}Desktop\\ Action \cite{cai2018desktop}\end{tabular} & 2018 & C & \begin{tabular}[c]{@{}c@{}}GoPro\\ Hero2\end{tabular} & H & $\sim$324k & 60 & 3 h & 6 & 1920$\times$1080 & \begin{tabular}[c]{@{}c@{}}act\\ msk\end{tabular} \\ \hline
\begin{tabular}[c]{@{}c@{}}Epic\\ Kitchens \cite{damen2018scaling}\end{tabular} & 2018 & C & GoPro & H & ~11.5M & - & 55 h & 32 & 1920$\times$1080 & act\\ \hline
FPHA \cite{garcia2018first} & 2018 & CD & \begin{tabular}[c]{@{}c@{}}Intel\\ RealSense\\ SR300\end{tabular} & S & \begin{tabular}[c]{@{}c@{}}$\sim$105k\\   (annotated)\end{tabular} & 1,175 & - & 6 & \begin{tabular}[c]{@{}c@{}}C:1920$\times$1080\\D:640$\times$480\end{tabular} & \begin{tabular}[c]{@{}c@{}}act\\ pos\end{tabular} \\ \hline
EYTH \cite{urooj2018analysis} & 2018 & C & - & - & \begin{tabular}[c]{@{}c@{}}1,290\\ (annotated)\end{tabular} & 3 & - & - & - & msk\\ \hline
EGTEA+ \cite{li2018eye} & 2018 & C & \begin{tabular}[c]{@{}c@{}}SMI wearable\\ eye-tracker\end{tabular} & H & \textgreater{}3M & 86 & $\sim$28 h & 32 & 1280$\times$960 & \begin{tabular}[c]{@{}c@{}}act\\ gaz\\ msk\end{tabular} \\ \hline
THU-READ \cite{tang2018multi} & 2018 & CD & \begin{tabular}[c]{@{}c@{}}Primesense\\ Carmine\end{tabular} & H & $\sim$343k & 1,920 & - & 8 & 640$\times$480 & \begin{tabular}[c]{@{}c@{}}act\\ msk\end{tabular} \\ \hline
EgoGesture \cite{zhang2018egogesture} & 2018 & CD & \begin{tabular}[c]{@{}c@{}}Intel\\ RealSense\\ SR300\end{tabular} & H & $\sim$3M & 24,161 & - & 50 & 640$\times$480 & gst\\ \hline
\begin{tabular}[c]{@{}c@{}}EgoDaily \cite{cruz2019my}\end{tabular} & 2019 & C & \begin{tabular}[c]{@{}c@{}}GoPro\\ Hero5\end{tabular} & - & $\sim$50k & 50 & - & 10 & 1920$\times$1080 & \begin{tabular}[c]{@{}c@{}}det\\ hid\end{tabular} \\ \hline
ANS SCI \cite{likitlersuang2019egocentric} & 2019 & C & \begin{tabular}[c]{@{}c@{}}GoPro\\ Hero4\end{tabular} & H & - & - & - & 17 & 1920$\times$1080 & \begin{tabular}[c]{@{}c@{}}det\\ int\end{tabular} \\ \hline
KBH \cite{wang2019recurrent} & 2019 & C & HTC Vive & H & \begin{tabular}[c]{@{}c@{}}$\sim$12.5k\\ (annotated)\end{tabular} & 161 & - & 50 & 230$\times$306 & msk\\ \hline
\end{tabular}
\end{center}
    \caption{List of available datasets with hand-based annotations in FPV. Image modality (Mode): Color (C); Depth (D); Color+Depth (CD). Camera location: Head (H); Chest (C); Shoulder (S). Annotation: actions/activities (\textbf{act}); hand presence and location (\textbf{det}); fingertip positions (\textbf{ftp}); gaze (\textbf{gaz}); grasp types (\textbf{grs}); hand gestures (\textbf{gst}); hand disambiguation (\textbf{hid}); hand-object interactions (\textbf{int}); hand segmentation masks (\textbf{msk}); object classes (\textbf{obj}); hand pose (\textbf{pos}).}
\label{tab_datasets}
\end{table*}

The importance that this field of research gained in recent years is clear when we look at the number of available datasets published since 2015 (Table \ref{tab_datasets}). Although the type of information and ground truth annotations made available by the authors is heterogeneous, it is possible to identify some sub-areas that are more recurrent than others. The vast majority of datasets provided hand segmentation masks \cite{li2013pixel,baraldi2014gesture,baraldi2015gesture,bambach2015lending,likitlersuang2016interaction,cai2018desktop,urooj2018analysis,li2018eye,tang2018multi,wang2019recurrent}, reflecting the high number of approaches proposed in this area and summarized in Section \ref{sec:localization}. However, the high number of datasets is counterbalanced by a relative low number of annotated frames, usually in the order of a few hundreds or thousands of images. To expedite the lenghty pixel-level annotation process and build larger datasets for hand segmentation, some authors proposed semi-automated techniques, for example based on Grabcut \cite{cai2018desktop, rother2004grabcut}.
Actions/activities \cite{pirsiavash2012detecting,wan2015mining,cai2018desktop,garcia2018first,li2018eye,tang2018multi} and hand gestures \cite{baraldi2014gesture,baraldi2015gesture,chang2016spatio,mohatta2017robust,wu2017yolse,xu2017hand,zhang2018egogesture} are other common information that were captured and annotated in many datasets. This large amount of data has been used by researchers for developing robust HCI applications that relied on hand gestures. Compared to the amount of hand segmentation masks, action/activities and hand gestures datasets are usually larger, since the annotation process is easier and faster than pixel-level segmentation.

The vast majority of datasets included color information recorded from head-mounted cameras. The head position is usually preferred over the chest or shoulders, since it is easier to focus on hand actions and manipulations whenever the camera wearer's is performing a specific activity. GoPro cameras were the most widely used devices for recording the videos, since they are specifically designed for egocentric POV and are readily available on the market. Few datasets, usually designed for hand pose estimation \cite{garcia2018first,yuan2017bighand2,xu2017hand}, hand gesture recognition \cite{chang2016spatio,zhang2018egogesture}, and action/activity recognition \cite{wan2015mining,garcia2018first,tang2018multi}, include depth or color and depth information. In most cases, videos were collected using Creative$^{\circledR}$ Senz3D\textsuperscript{TM} or Intel$^{\circledR}$ RealSense\textsuperscript{TM}  SR300 depth sensors, as these devices were small and lightweight. Moreover, these cameras were preferred over other depth sensors (e.g., Microsoft$^{\circledR}$ Kinect\textsuperscript{TM}) because they were originally developed for natural user interface that made them more suitable for studying hand movements in the short range (i.e., up to 1 m of distance from the camera).

Although FPV is gaining a lot of interest for developing healthcare applications for remote monitoring of patients living in the community, only one dataset (i.e., the ANS-SCI dataset \cite{likitlersuang2019egocentric}) included videos from people with clinical conditions such as cSCI. This lack of available data is mainly due to ethical constraints that make it harder to share videos and images collected from people with diseases or clinical conditions. In the next few years researchers should try -- within the ethical and privacy constraints -- to build and share datasets for healthcare applications including videos collected from patients. This will benefit the robustness of the hand-based approaches in FPV against the inter-group and within group variability that can be encountered in many clinical conditions.


\section{Conclusion}\label{sub:conclusion}

In this paper we showed how hand-related information can be retrieved and used in egocentric vision. We summarized the existing literature into three macro-areas, identifying the most prominent approaches for hand localization (e.g., hand detection, segmentation, pose estimation, etc.), interpretation (grasp analysis, gesture recognition, action and activity recognition), as well as the FPV applications for building real-world solutions. We believe that a comprehensive taxonomy and an updated framework of hand-based methods in FPV may serve as guidelines for the novel approaches proposed in this field by helping to identify gaps and standardize terminology.

One of the main factors that promoted the development of FPV approaches for studying the hands is the availability of wearable action cameras and AR/VR systems. However, we also showed how the use of depth sensors, although not specifically developed for wearable applications, has been exploited by many authors, in order to improve the robustness of hand localization. We believe that the possibility to develop miniaturized wearable depth sensors may further boost the research in this area and the development of novel solutions, since a combination of color and depth information can improve the performance of several hand-based methods in FPV. In particular, these advantages can be exploited in settings that involve short--range observations (i.e., less than 1 m) and indoor environments, which are often encountered when analyzing the hands in FPV.

From this survey it is clear how the hand localization step plays a vital role in any processing pipeline, as a good localization is necessary condition for hand-based higher inference, such as gesture or action recognition. This importance has motivated the extensive research conducted in the past 10 years, especially in sub-areas like hand detection and segmentation. The importance of hand localization methods may also be seen in those approaches where the hands play an auxiliary role, such as activity recognition. In fact, the position of the hands can be used to build attention-based classifiers, where more weight is given to the manipulation area.

Like other computer vision fields, the advent of deep learning has had a great impact on this area, by boosting the performance of several localization and interpretation approaches, as well as optimizing the number of steps required to pursue a certain objective (see the hand identification example -- Section \ref{subsec:handIdent}). Hand detection is the localization sub-area that has seen the largest improvements, especially thanks to the availability of object detection networks retrained on large datasets. Other sub-areas, such as hand segmentation and pose estimation, perhaps will see larger improvements in the next few years, especially if the amount of available data annotations grows. Recurrent models, 3DCNN, and the availability of large datasets (e.g., Epic Kitchens, EGTEA+, etc.) have helped pushing the state of the art of action and activity recognition, considering that the combination of temporal and appearance information was demonstrated to be crucial for these tasks. In the near future, efforts should be made in improving methods for the recognition of larger classes of unscripted ADLs, which would benefit the development of applications such as AAL.

As this field of research is still growing, we will see novel applications and improvement of the existing ones. The impact of hand-based methods in egocentric vision is clear from the development of applications that relied on HCI and HRI. The importance of the hands as our primary means of interaction with the world around us is currently exploited by VR and AR systems, and the position of the wearable camera offers tremendous advantages for assessing upper limb function remotely and supporting older adults in ADLs. This will translate in the availability of rich information captured in natural environments, with the possibility to improve assessment and diagnosis, provide new interaction modalities, and enable personalized feedback on tasks and behaviours.

\section*{Acknowledgments}
This work was supported in part by the Craig H. Neilsen Foundation (542675). The authors would also like to thank Gustavo Balbinot, Guijin Li, and Mehdy Dousty for the helpful discussions and feedback.


%

\bibliographystyle{unsrt}
\bibliography{references}

%

\begin{IEEEbiography}[{\includegraphics[width=1in,height=1.25in,clip,keepaspectratio]{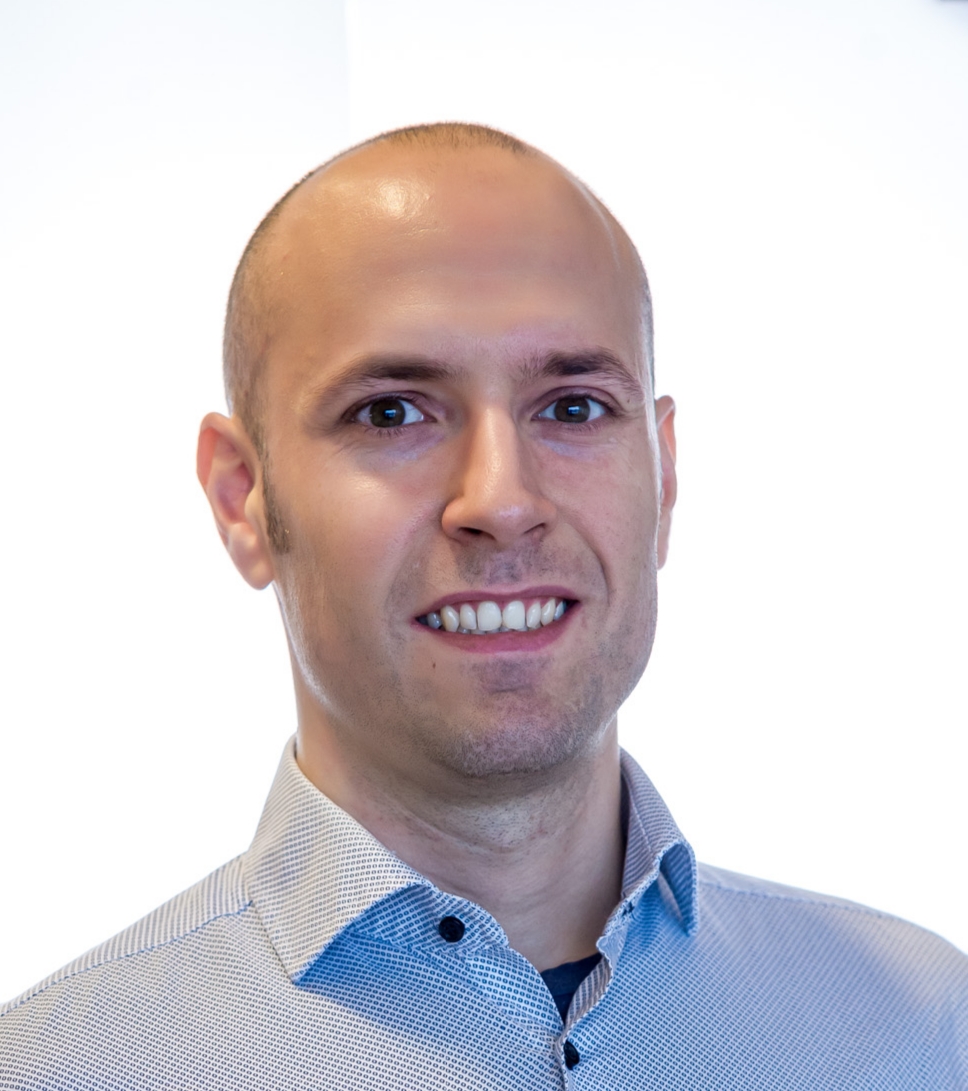}}]{Andrea Bandini}
Andrea Bandini (M'16) received his Master's degree in Biomedical Engineering from the University of Firenze (Italy) in 2012, and the PhD the Bioengineering from the University of Bologna (Italy) in 2016. He has been a postdoctoral research fellow at KITE - University Health Network (Toronto, Canada) since September 2016. His research aims at developing intelligent tools for remote assessment and rehabilitation of motor signs associated with neurological disorders (spinal cord injury, stroke, amyotrophic lateral sclerosis, and Parkinson’s disease), by using computer vision and machine learning techniques.

\end{IEEEbiography}

\begin{IEEEbiography}[{\includegraphics[width=1in,height=1.25in,clip,keepaspectratio]{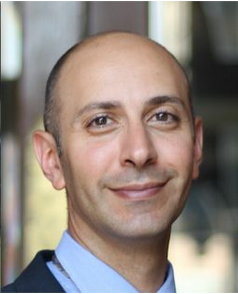}}]{José~Zariffa}
José Zariffa (M’01, SM’18) received the Ph.D. degree in 2009 from the University of Toronto's Department of Electrical and Computer Engineering and the Institute of Biomaterials and Biomedical Engineering. He later completed post-doctoral fellowships at the International Collaboration On Repair Discoveries (ICORD) at the University of British Columbia in Vancouver, Canada, and at the Toronto Rehabilitation Institute – University Health Network in Toronto, Canada. He is a currently a Scientist at KITE - Toronto Rehabilitation Institute - University Health Network and an Associate Professor at the Institute of Biomaterials and Biomedical Engineering at the University of Toronto in Toronto, Canada. His research interests include technology for upper limb rehabilitation after spinal cord injury, neural prostheses, and interfaces with the peripheral nervous system. Dr. Zariffa is the recipient of an Ontario Early Researcher Award.

\end{IEEEbiography}

\end{document}